%% file: acl_latex.tex
\documentclass[11pt]{article}

\usepackage[preprint]{acl}

\usepackage{times}
\usepackage{latexsym}
\usepackage{booktabs}
\usepackage{multirow}
\usepackage{caption}

\usepackage[T1]{fontenc}

\usepackage[utf8]{inputenc}
\usepackage{microtype}
\usepackage{inconsolata}
\usepackage{graphicx}
\usepackage{amsmath}
\usepackage{mathtools}

\title{What Shapes Brain Alignment of Large Language Models during Dialogue? Insights from Language Production and Comprehension}

\author{Yuko Nakagi,  Yuya Chiba,  Yasue Kishino \\
        NTT, Inc. \\
   \texttt{yuko.nakagi@ntt.com}
        }

\begin{document}
\maketitle

\input{sec/0_abstract} 
\input{sec/1_introduction}
\input{sec/2_relatedwork}
\input{sec/3_method}
\input{sec/4_results}
\input{sec/5_discussion}
\input{sec/6_limitations}

\input{sec/7_ethics}
\bibliography{custom}
\clearpage
\input{sec/appendix}

\end{document}

%% file: sec/0_abstract.tex
\begin{abstract}
\label{sec:abstract}

Brain alignment of large language models (LLMs) offers a way to investigate representational similarities between LLMs and the human brain during language processing.
While previous studies have mainly focused on passive language comprehension, little is known about such alignment during dialogue, in which speakers engage in both language production and comprehension.
In this study, using fMRI data collected during dialogue, we examine brain alignment during language production and comprehension across LLM layers and brain regions.
We assess its association with the intrinsic dimension (ID) of LLM representations and the contributions of several interpretable linguistic and communicative features.
Our results reveal distinct patterns of brain alignment between production and comprehension, as well as strong associations with ID in brain regions associated with each processing mode. Furthermore, we show that the contributions of linguistic and communicative features also vary across processing modes, LLM layers, and brain regions.

\end{abstract}

%% file: sec/1_introduction.tex
\section{Introduction}

Elucidating the neural mechanisms underlying human language processing and communication is a major goal in neuroscience.
Recent studies have shown that latent representations extracted from language models can effectively predict brain activity, as they capture richer linguistic information than hand-crafted linguistic features.
This correspondence between brain and large language models (LLMs), referred to as \textbf{brain alignment}, has been widely studied ~\cite{Jain-nips2018,Schrimpf2021-vr,Goldstein2022-yw} and this alignment is strongest for intermediate layers of LLMs ~\cite{NEURIPS2019_749a8e6c,Caucheteux2022-hl,Caucheteux2022-yb,antonello2023scaling,Mischler2024-oi}.
More recently, increasing attention has been paid to identifying the model-derived factors~\citep{Cheng2026-yc,Proietti2026-icml} and stimulus-derived linguistic factors ~\citep{caucheteux2021disentangling,Oota_2023,oota-etal-2024-speech,alkhamissi-etal-2025-language} underlying this alignment.
These studies have provided insights into the representational properties that contribute to brain alignment.

Previous studies have primarily investigated brain alignment and its underlying representational properties during passive language comprehension, such as reading or listening to narratives.
However, everyday language use is not limited to passive comprehension.
One of the most common forms of language-mediated communication is conversation. 
Unlike unidirectional language comprehension, conversation is inherently interactive and involves both language production and comprehension, with individuals alternating between producing their own utterances and comprehending those of their interlocutors.
Each utterance conveys multiple types of information, including lexical, compositional, and semantic information as well as communicative intentions that coordinate the ongoing dialogue.
Recently, a growing number of studies have examined brain alignment during language production and comprehension in dialogue~\cite{Zada2024-to,Goldstein2025-yb,Yamashita2025-xx,Cai2025-kw,Zada2025-ep}. 
Yet, how brain alignment varies across LLM layers and brain regions during production and comprehension remains insufficiently understood.
Furthermore, it remains unclear which dialogue-related representational properties contribute to brain alignment.

Our study aims to provide a detailed analysis of brain alignment of LLMs during dialogue and to investigate the representational properties underlying this alignment.
First, using fMRI data collected during dialogue and latent representations extracted from multiple LLMs, we investigate brain alignment during production and comprehension across brain regions and LLM layers. 
Next, we examine the extent to which model-derived and dialogue-derived factors account for this alignment.
Specifically, we investigate the relationship between brain alignment and intrinsic dimension (ID), a model-derived measure of representational complexity previously identified as one of the properties most strongly associated with brain alignment~\cite{Cheng2026-yc}.
However, ID alone does not reveal which specific dialogue-derived linguistic and communicative features contribute to brain alignment.
We therefore investigate the contributions of four interpretable dialogue-derived features, ranging from surface-level linguistic properties to communicative intentions.
Through these analyses, we show that (1) brain alignment during dialogue varies systematically across processing modes (production vs.\ comprehension), brain regions, and LLM layers, and that (2) this alignment is strongly associated with ID, particularly in brain regions closely associated with each processing mode, while the contributions of interpretable dialogue-derived features vary across processing modes, LLM layers, and brain regions.

To the best of our knowledge, this is the first study to decompose brain alignment of LLMs during dialogue in terms of representational features at multiple levels. 
Our study extends the interpretability of brain alignment beyond the brain's language network to higher-order cognitive regions involved in dialogue.

\label{sec:introduction}

%% file: sec/2_relatedwork.tex
\section{Related Work}
\label{sec:related_work}

To examine brain alignment, researchers have used brain encoding models ~\cite{Naselaris2011-wm,NISHIMOTO20111641,Huth2012-bh} to predict brain activity from latent representations extracted from LLMs.
A growing body of research has investigated this correspondence, primarily focusing on the relationship between LLM representations and brain activity during language comprehension~\cite{NEURIPS2019_749a8e6c,Caucheteux2022-hl,Caucheteux2022-yb,antonello2023scaling,Mischler2024-oi,NEURIPS2024_f96839fc,Gao2025-pa,Goldstein2025-gx}.
These studies have consistently shown that latent representations from intermediate LLM layers are particularly effective at predicting brain activity.
Several factors have been proposed to account for this alignment, including next-word prediction capability, as reflected in surprisal~\cite{goldstein2022shared}, linguistic information at the lexical, syntactic, and semantic levels~\cite{caucheteux2021disentangling,Oota_2023,alkhamissi-etal-2025-language,he2025far}, and representational complexity~\cite{Cheng2026-yc}.
These factors have contributed to both understanding brain alignment and improving its alignment.

Recent studies have extended this line of research to dialogue. 
Previous research has identified shared and unique linguistic representations between production and comprehension~\cite{Zada2024-to}.
These shared and unique representations have also been characterized across different context lengths~\cite{Yamashita2025-xx}.
Another study demonstrated that both acoustic and textual representations extracted from speech language models correspond to brain activity~\cite{Goldstein2025-yb}.
However, fine-grained analyses of brain alignment across brain regions and LLM layers during dialogue remain limited~\citep{Cai2025-kw}, particularly at the whole-brain level and for more recent models. 
Moreover, whether brain alignment during dialogue varies with the representational properties of LLMs remains unexplored.

%% file: sec/3_method.tex
\section{Dataset and Models}
\label{sec:methods}

\subsection{fMRI Dataset}
\label{sec:methods-fmridataset}

This study uses the Natural Dialogue fMRI Dataset~\cite{Yamashita2025-xx}.
This dataset provides brain activity recorded while participants engaged in dialogue on various topics inside a 3T functional magnetic resonance imaging (fMRI) scanner.
All participants were native Japanese speakers.
We analyze data from the 7 participants (sub-OSU01 to sub-OSU07).

The dialogue topics covered a broad range of everyday-life subjects, including self-introduction, family, and moments of happiness.
In each run, participants engaged in a dialogue on a given topic with an interlocutor outside the MRI scanner, freely expressing their thoughts and emotions in real time. 
Each run lasted 7 min 10 s.
We use the transcriptions of both speakers' utterances in the dataset, which are temporally aligned with the fMRI repetition time (TR$=$1s).
To align the dialogue with the fMRI data, we treat each one-second segment as a unit of analysis.

We follow the session partitioning provided in the dataset to construct the training and test sets.
To standardize the number of sessions to three across participants, we exclude sessions consisting of four or fewer runs.
We also exclude two runs from sub-OSU03 due to missing transcription data.
After these exclusions, each participant has three sessions, each consisting of 7--11 runs.
We then perform three-fold leave-one-session-out cross-validation, using one session as the test set and the remaining two sessions as the training set in each fold.
All prediction accuracies reported in this paper are evaluated on the held-out test sets.

We conduct region-of-interest (ROI) analyses using the ROIs provided with the dataset, which are defined based on the anatomical parcellation of the Destrieux Atlas~\cite{destrieux2010automatic}.
We focus on eight regions previously implicated in information encoding during language production and comprehension: the anterior transverse temporal gyrus (A1), superior temporal gyrus (STG), superior temporal sulcus (STS), middle temporal gyrus (MTG), angular gyrus (AG), middle frontal gyrus (MFG), inferior frontal gyrus (IFG), and sensorimotor areas (SM)~\citep{binder1997human,Silbert_2014,chai2016functional, Goldstein2025-yb, Cai2025-kw, Zada2025-ep}. 
Further details on the ROIs are provided in Section~\ref{appendix:additional_methods:fmri_data}.

\subsection{Large Language Models}
\label{sec:methods-llm}

Because this study analyzes a Japanese dataset, we use three LLMs tuned for Japanese: Llama-3.1-Swallow, Gemma-2-Swallow, GPT-OSS-Swallow~\citep{Fujii:COLM2024,Okazaki:COLM2024,ma2025building}.
These models have between 8–21B parameters.
Although their architectures differ, all three are causal decoder-only Transformer-based language models trained on large-scale text corpora using next-token prediction.
While many previous studies have investigated brain alignment using pretrained models, we use post-trained models with conversational capabilities to examine brain alignment during dialogue.
Further details on the models are provided in Section~\ref{appendix:additional_methods:llms}.
Hereafter, we refer to Llama-3.1-Swallow, Gemma-2-Swallow, and GPT-OSS-Swallow as Llama-3.1, Gemma-2, and GPT-OSS, respectively.
We use Llama-3.1 for the main analysis and present the results for Gemma-2 and GPT-OSS in the Appendix.

\section{Methods}

\subsection{Extraction of Latent Representations}
\label{sec:methods-extraction-llm-repsentation}

We extract latent representations from each LLM layer for utterances produced by the participant ($P$) and the interlocutor ($I$).
To incorporate dialogue context, at each time point $t$, we input the target speaker's current utterance together with the preceding dialogue from both speakers within a context window of length $l$.
Formally, for a target speaker $s \in \{P,I\}$, we define $u_t^s$ as the transcription of speech produced during the 1TR segment $x_t^s$ at time point $t$.
We define the context window spanning time points $t-l$ to $t-1$ as $x_{t-1}^{(l)} \coloneq x_{t-l:t-1}$.
For each speaker, we concatenate consecutive utterance segments within $x_{t-1}^{(l)}$ and arrange the resulting utterances in chronological order based on their onset times, yielding the dialogue context $c_{t,l}$.

We then define the latent representations of the participant and the interlocutor at time point $t$ with context length $l$, denoted by $h_{t,l}^{P}$ and $h_{t,l}^{I}$, respectively, as follows:
\[
h_{t, l}^{P}
=
\pi\left([c_{t,l},\,u_t^P]\right)
\]
\[
h_{t, l}^{I}
=
\pi\left([c_{t,l},\,u_t^P,\,u_t^I]\right)
\]
Here, $\pi$ denotes the function that extracts the latent representation of the final token in the input sequence from each LLM layer.
We extract these representations for context lengths $l \in \{1,4,16,32\}$.

\subsection{Brain Alignment of LLMs}
\label{sec:methods-alignment}
To investigate the brain alignment of LLMs during dialogue, we construct voxel-wise brain encoding models following the approach described by \citet{Yamashita2025-xx}.
For each participant and LLM layer, we construct separate encoding models for production and comprehension.
We use $h_{t,l}^{P}$ and $h_{t,l}^{I}$ for production and comprehension, respectively.
To account for the hemodynamic delay in the BOLD signal, we consider time delays ranging from 2 to 9 s.
To reduce the influence of head motion, we remove head-motion-related components from the fMRI activity before fitting the encoding models (see Appendix~\ref{appendix:additional_methods:headmotion} for details).
We estimate the model weights using $\ell_2$-regularized linear regression to map latent representations to fMRI activity in the training data.
The regularization parameter is selected separately for each voxel using five-fold cross-validation on the training data, using candidate values from $10^{-2}$ to $10^7$.
We then apply the estimated weights to the LLM representations of the test data.
Prediction accuracy is evaluated using the Pearson correlation coefficient between the observed and predicted fMRI activity in the test data.
Statistical significance is assessed using a one-sided permutation test.
We generate a null distribution by randomly shuffling 20-TR blocks of the test data 1,000 times and computing the correlation coefficient for each permutation.
The significance threshold is set to $P < 0.05$, with false discovery rate (FDR) correction for multiple comparisons. 
We implement the brain encoding models using the \texttt{himalaya} library~\cite{la2022feature}\footnote{\url{https://github.com/gallantlab/himalaya}}.

\subsection{Comparison between Brain Alignment and Representational Complexity}
\label{sec:methods-comp-abstmeaning}
To examine whether brain alignment during production and comprehension is associated with the complexity of LLM representations, we analyze ID, a strong predictor of brain alignment~\cite{Cheng2026-yc}.
ID estimates the number of dimensions underlying the nonlinear structure of the representation space and has been shown to be associated with the construction of syntactic and semantic features across LLM layers~\citep{cheng2025emergence,lee2025geometric,baroni2026tracing}.
We estimate ID from the layerwise latent representations extracted in Section~\ref{sec:methods-extraction-llm-repsentation} using the Generalized Ratios Intrinsic Dimension Estimator (GRIDE)~\citep{Denti2022}.
We compute Spearman rank correlations between ID and brain alignment across layers and compare these correlations between production and comprehension and across ROIs.
Estimation details are provided in Section~\ref{appendix:additional_methods:IDs}.

\begin{figure*}[t]
\centering
\includegraphics[width=0.95\linewidth]{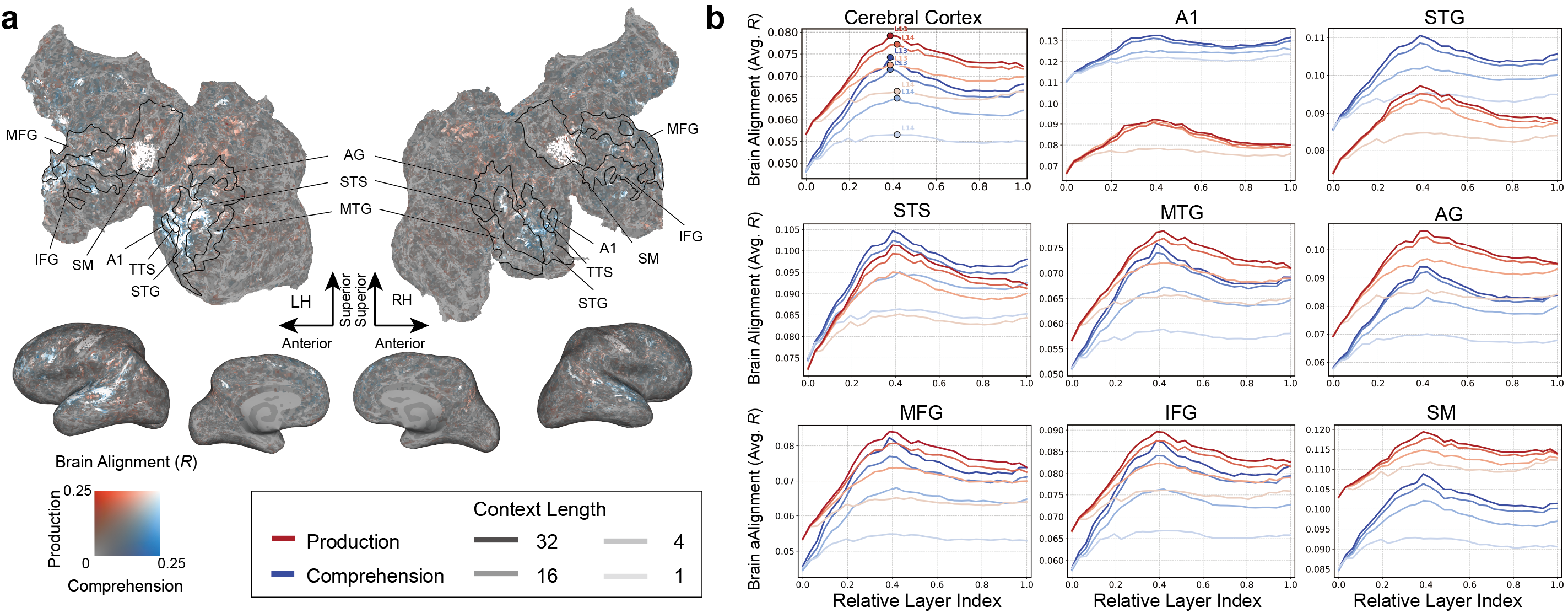}
\caption{\textbf{Brain Alignment of LLMs during Dialogue.} 
\textbf{a} Brain alignment for production and comprehension in one participant (sub-OSU01), using layer 13 of Llama-3.1, projected onto the inflated (lateral and medial views on both sides) and flattened cortical surfaces (occipital areas at the center) of both hemispheres. Red and blue indicate production and comprehension, respectively. Voxels significant in either production or comprehension are colored (all colored voxels $P<0.05$, FDR corrected). 
\textbf{b} Mean brain alignment across the cerebral cortex (top left) and within each ROI (remaining panels) for production and comprehension across Llama-3.1 layers, averaged across participants. Color intensity indicates context length, and circles indicate the peak layer for each condition. 
}
\label{fig:prod_vs_comp}
\end{figure*}

\subsection{Removal of Interpretable Features Related to Dialogue}
\label{sec:methods-comp-commint}
The ID-based analysis in Section~\ref{sec:methods-comp-abstmeaning} alone cannot reveal how specific dialogue-related features contribute to brain alignment.
One way to assess the contribution of these features is to remove information about target features from LLM representations and examine the effects of this removal on brain alignment~\cite{Toneva2022-nd,Oota_2023,oota-etal-2024-speech}.
For feature removal, we use LEAst-squares Concept Erasure (LEACE)~\cite{NEURIPS2023_d066d21c}, a method that removes linearly decodable information about a target concept while minimizing changes to the original representations.
We use concept scrubbing~\cite{NEURIPS2023_d066d21c}, which applies LEACE sequentially across all layers.
% Starting from the first layer, we apply LEACE, pass the resulting representations to the next layer, and repeat this procedure through the final layer.

We consider four types of features, ranging from surface-level linguistic properties to communicative intentions, as target concepts for LEACE and apply concept scrubbing to all three LLMs.
For surface-level features, we use bunsetsu\footnote{A bunsetsu is a natural minimal unit of a Japanese sentence, consisting of either an independent word or an independent word followed by one or more dependent words. It roughly corresponds to a basic phrase in English.} count (BC), a low-level feature reflecting utterance length, and lexicon (L), which captures lexical information.
For syntactic information, we use part-of-speech (POS) features, and for communicative intentions, we use dialogue acts (DA).
Concept scrubbing removes linearly decodable information about a target feature using labels corresponding to that feature.
We therefore construct one-hot or multi-hot feature vectors from each speaker's TR-level text segments and concatenate them into feature matrices with 6--1,000 dimensions.
Details of each feature are provided in Section~\ref{appendix:additional_methods:interpretable_features}.

We fit LEACE on the training data and construct brain encoding models from the resulting LLM representations following Section~\ref{sec:methods-alignment}.
We assess the effects of feature removal on brain alignment by comparing prediction accuracy before and after removal.
We also compare probing performance before and after feature removal to assess how effectively the target features are removed from LLM representations (see Section~\ref{sec:app-probing-interp-feature}).

%% file: sec/4_results.tex
\section{Results}
\label{sec:results}

\begin{figure*}[t]
\centering
\includegraphics[width=0.95\linewidth]{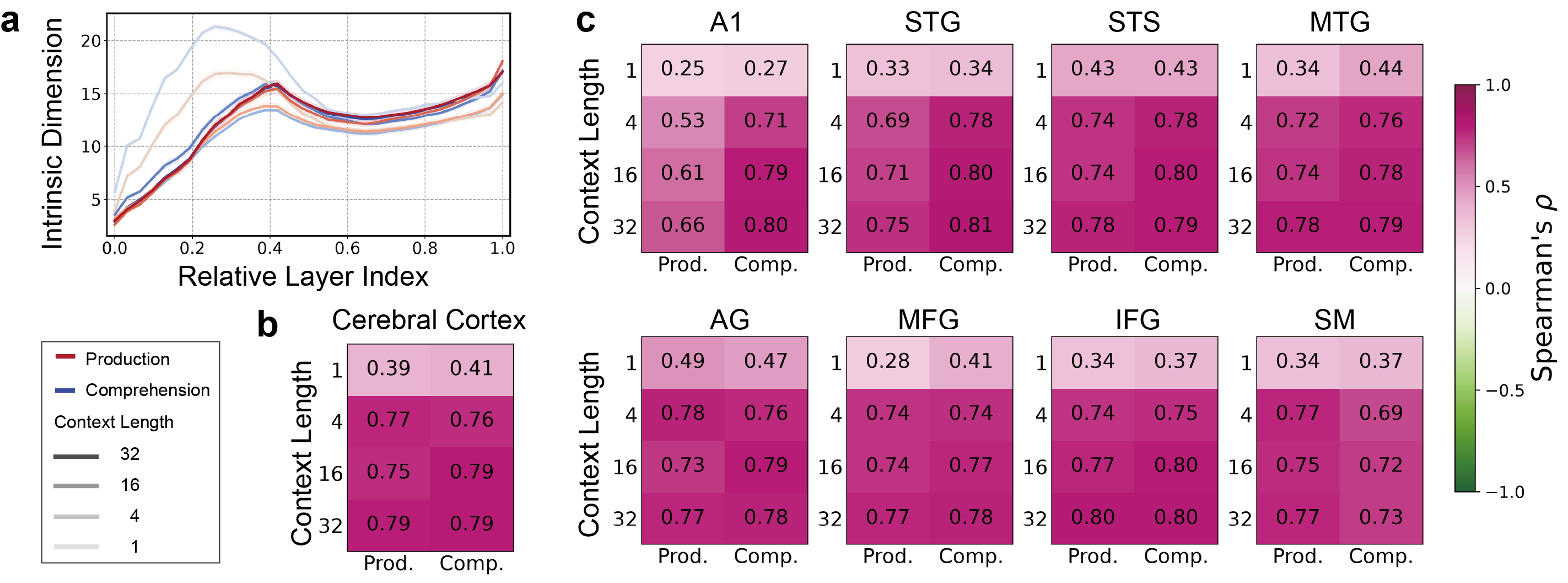}
\caption{\textbf{Relationship between LLM representational complexity and brain alignment during dialogue.} \textbf{a} Layerwise ID of Llama-3.1 representations for production and comprehension, averaged across participants' dialogue. Shaded regions indicate the standard error of the mean. Red and blue indicate production and comprehension, respectively, and color intensity indicates context length. \textbf{b} Spearman rank correlations between layerwise normalized ID and layerwise whole-cortex-averaged brain alignment for production and comprehension at each context length. Correlation coefficients are averaged across all participants and LLMs in Fisher $z$ space. 
\textbf{c} Spearman rank correlations between layerwise normalized ID and layerwise ROI-averaged brain alignment.}
\label{fig:ID_brain-alignment}
\end{figure*}

\subsection{Brain Alignment of LLMs during Dialogue}
\label{sec:results-brain-alignment}

First, we examine whether intermediate-layer representations of LLMs can predict brain activity for production and comprehension.
Given previous findings that brain alignment generally peaks at intermediate layers, we focus on an intermediate layer.
Figure~\ref{fig:prod_vs_comp}a shows voxel-wise whole brain alignment for Layer 13 of Llama-3.1.
The encoding models accurately predict brain activity across a wide range of cortical regions, including association areas.
Specifically, for production, the models accurately predict activity in auditory, sensorimotor, and parietal regions, whereas for comprehension, they predict activity in auditory, sensorimotor, and frontal regions, consistent with previous studies~\cite{Zada2024-to,Goldstein2025-yb,Zada2025-ep}.
These results demonstrate that LLM representations significantly predict brain activity and that the spatial patterns of brain alignment differ between production and comprehension across brain regions.

Next, we examine brain alignment across LLM layers for production and comprehension.
The top-left panel of Figure~\ref{fig:prod_vs_comp}b shows whole-cortex-averaged brain alignment across Llama-3.1 layers for production and comprehension.
The layerwise trajectories of brain alignment are similar for production and comprehension, with alignment gradually increasing from early layers and peaking at intermediate layers.
Brain alignment also increases with the length of the dialogue context provided to the LLM.
These results indicate that intermediate-layer LLM representations show particularly strong brain alignment when longer dialogue context is provided.

Third, we examine layerwise brain alignment across ROIs for production and comprehension.
The remaining panels of Figure~\ref{fig:prod_vs_comp}b show ROI-averaged brain alignment across Llama-3.1 layers for production and comprehension.
For production, alignment remains consistently high across layers in SM and AG, whereas for comprehension, it remains consistently high in A1 and STG.
The layerwise trajectories of brain alignment for production and comprehension are similar up to intermediate layers across ROIs; however, in A1, brain alignment during comprehension tends to increase again toward the output layers (relative layer index: 0.9--1.0).
These results suggest that the relationship in the layerwise trajectories of brain alignment between production and comprehension varies across ROIs, particularly near the output layers.
Results for individual participants and all LLMs are provided in Section~\ref{sec:app-results-brain-alignment}.

\subsection{Representational Properties and Brain Alignment during Dialogue}
To investigate what changes in LLM representations themselves may account for the observed changes in brain alignment, we first quantify the ID of LLM representations and examine its relationship with brain alignment.
ID has been shown to be associated with brain alignment during language comprehension~\citep{Cheng2026-yc}.
In this study, we examine whether this relationship also holds for brain alignment during dialogue.

Figure~\ref{fig:ID_brain-alignment}a shows the layerwise ID of Llama-3.1.
As observed for brain alignment, ID exhibits similar layerwise trajectories for production and comprehension (see Figure~\ref{fig:App6_id_plot} for the results of the other LLMs).
Figure~\ref{fig:ID_brain-alignment}b shows the Spearman rank correlation coefficients between layerwise ID and layerwise whole-cortex-averaged brain alignment, revealing strong correlations for both production and comprehension when the context length is $\geq 4$.
These results suggest that ID is associated with brain alignment not only during passive language comprehension but also during dialogue.

Previous work has shown that the strength of the association between layerwise brain alignment and layerwise ID varies across voxels~\citep{Cheng2026-yc}.
However, how this relationship differs across brain regions during dialogue remains unclear.
To address this, we examine the Spearman rank correlations between layerwise ID and layerwise ROI-averaged brain alignment for each ROI, as shown in Figure~\ref{fig:ID_brain-alignment}c.
In auditory regions, brain alignment tends to correlate more strongly with ID for comprehension than for production, whereas in SM, it tends to correlate more strongly with ID for production than for comprehension.
In language-related regions (AG, MFG, and IFG), the correlations are largely comparable between production and comprehension.
In particular, brain alignment in A1 correlates more strongly with ID for comprehension than for production, consistent with the differences in layerwise trajectories of brain alignment between production and comprehension observed in these regions in Figure~\ref{fig:prod_vs_comp}c.
These results suggest that brain alignment in ROIs predominantly involved in production or comprehension is more strongly associated with ID for the corresponding process.

\subsection{Effects of Interpretable Features Related to Dialogue}
\label{sec:results-Interp_feature}
Finally, we investigate the extent to which the four interpretable dialogue-related features defined in Section~\ref{sec:methods-comp-commint} contribute to the brain alignment of LLMs during dialogue.
We focus on a context length of 32, which yields the highest brain alignment, as shown in Figure~\ref{fig:prod_vs_comp}b.

\begin{table}[t]
\caption{\textbf{Mean Percentage Decrease in Brain Alignment after Removing Interpretable Features.} Percentage decrease ($\%$) in whole-cortex-averaged brain alignment of Llama-3.1 after removing each of the four feature types. Results are averaged across participants and shown for layers 1, 13, 26, and 32 for both production (Prod) and comprehension (Comp).}
\centering
\small
\begin{tabular}{llrrrr}
\toprule
 & & L1 & L13 & L26 & L32 \\
\midrule
\multirow{4}{*}{Prod}
& BC  & 32.21 & 7.81  & 12.94 & 12.12 \\
& L  & \textbf{85.89} & \textbf{41.57} & \textbf{49.15} & \textbf{45.95} \\
& POS & 71.57 & 28.73 & 38.06 & 36.75 \\
& DA  & 68.42 & 25.82 & 34.39 & 32.85 \\
\midrule
\multirow{4}{*}{Comp}
& BC  & 27.63 & 8.36  & 9.06  & 7.03 \\
& L  & \textbf{78.60} & \textbf{33.53} & \textbf{43.71} & \textbf{38.40} \\
& POS & 74.52 & 30.22 & 38.89 & 33.05 \\
& DA  & 75.90 & 31.46 & 43.12 & 37.48 \\
\bottomrule
\end{tabular}
\label{tab:llama_layer_results}
\end{table}

\begin{figure*}[t]
\centering
\includegraphics[width=0.9\linewidth]{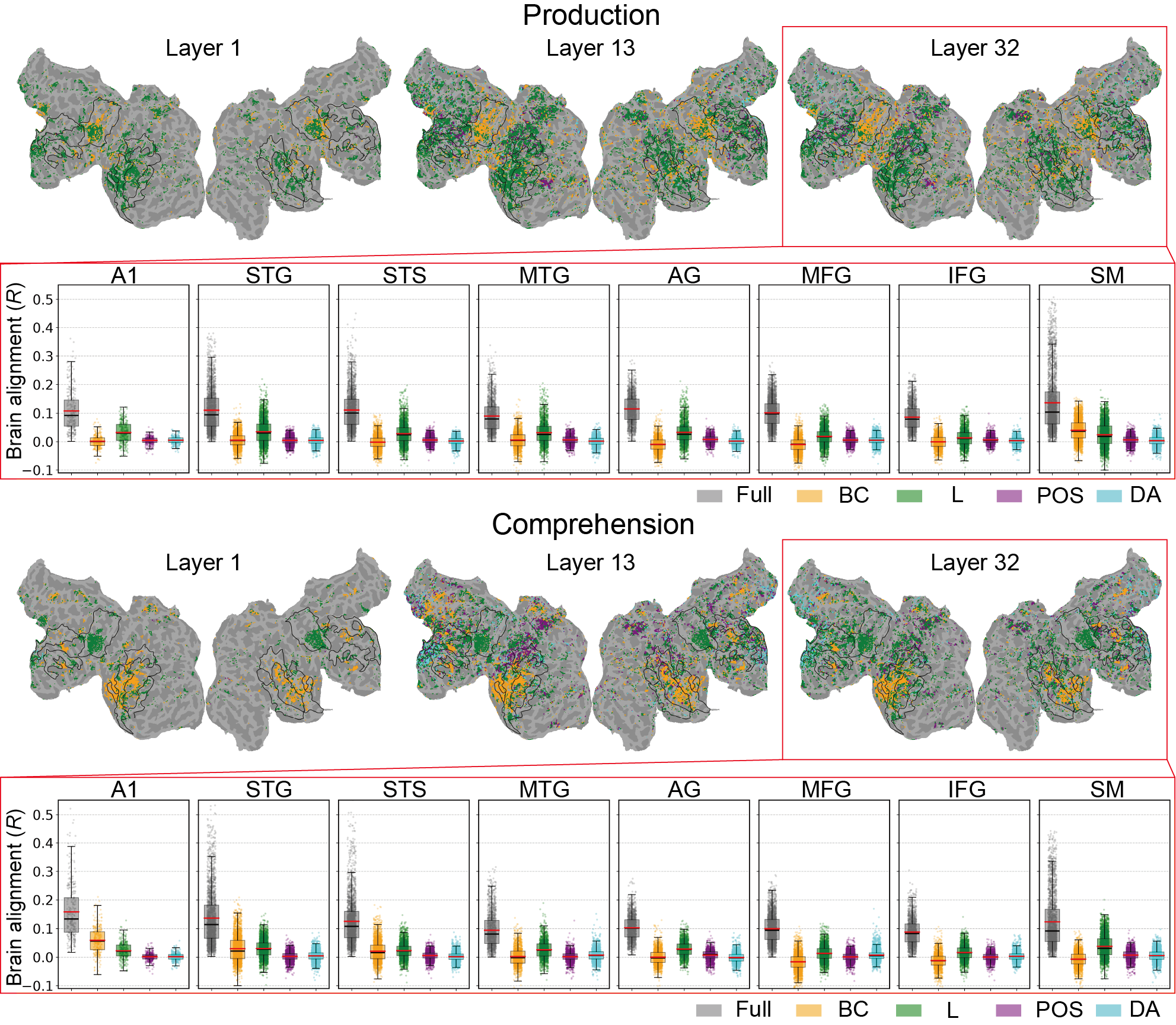}
\caption{\textbf{Effects of Interpretable Features Related to Dialogue.} \textit{Flatmaps}: For Layers 1, 13, and 32 of Llama-3.1, the feature whose removal caused the largest decrease in brain alignment at each voxel is mapped onto the cerebral cortex of sub-OSU01. Only voxels with significant brain alignment ($P < 0.05$, FDR corrected) of at least 0.1 in the full model are colored. \textit{Box plots}: The unique contributions of each of the four features to brain alignment in each ROI at the final layer of Llama-3.1 for sub-OSU01. Boxes indicate the median and interquartile range, red lines indicate the mean, and points represent individual voxels. Gray indicates brain alignment in the full model, while yellow, green, purple, and blue indicate the unique contributions of BC, L, POS, and DA, respectively.}
\label{fig:effect-interp-feat}
\end{figure*}

First, we compute brain alignment using LLM representations from which each feature is individually removed, and assess the contribution of each feature to brain alignment.
Table~\ref{tab:llama_layer_results} shows the percentage decrease in whole-cortex-averaged brain alignment at four layers of Llama-3.1.
In both production and comprehension, L makes the largest contribution to brain alignment across layers, with this tendency being particularly pronounced for production.
This suggests that concrete linguistic information contributes more strongly to brain alignment than lower-level or more abstract information represented in LLM representations.
However, these four features are not independent of one another. 
For example, the types of words used in dialogue may vary with utterance length, and utterances associated with certain DAs, such as backchannels, tend to be short.
It therefore remains unclear to what extent each feature makes a unique contribution to brain alignment.

To address this, we examine the contribution of each feature to the brain alignment of LLMs after cumulatively removing the effects of lower-level features.
Figure~\ref{fig:effect-interp-feat} shows the results for representative layers at which changes in feature contributions are clearly observable.
As shown in the figure, the four types of interpretable features represented at a given LLM layer uniquely capture brain activity, with their contributions varying across processing modes and brain regions during dialogue.
The flatmap results show that BC is dominant in regions around SM for production and in auditory areas such as A1 and STG for comprehension.
At the ROI level, BC makes a significantly greater unique contribution than the other features in A1 during comprehension for five participants ($P < 0.001$, two-sided Wilcoxon signed-rank tests with Holm correction).
This suggests that information about utterance duration represented in LLMs captures brain signals related to articulatory movements and lower-level information processing.
In addition, L is dominant across a wide range of regions, including temporal, parietal, and frontal areas, in both modes.
In particular, we observe the dominance of L (and POS in some participants~\footnote{As shown in Figures~\ref{fig:App7_interp_feat_llama_prod},~\ref{fig:App8_interp_feat_llama_comp},~\ref{fig:App11_interp_feat_gemma_prod},~\ref{fig:App12_interp_feat_gemma_comp},~\ref{fig:App15_interp_feat_gptoss_prod}, and~\ref{fig:App16_interp_feat_gptoss_comp}, POS is dominant in the temporoparietal junction for sub-OSU01, sub-OSU03, and sub-OSU07 under some conditions.}) in the temporoparietal junction, including AG.
At the ROI level, L makes a significantly greater unique contribution than the other features in AG for all participants ($P < 0.001$, two-sided Wilcoxon signed-rank tests with Holm correction).
This suggests that lexical and syntactic information about utterances represented in LLMs captures brain activity related to word and semantic processing.
In contrast, the flatmap results show that POS and DA are dominant in regions surrounding those where BC and L are dominant.
Especially, DA tends to be dominant in higher-order brain regions, including frontal cortex, during both production and comprehension.
This suggests that DA, which represents communicative intention and is the most abstract of the four features, captures brain activity involved in relatively higher-order cognitive processing.
Results for all participants and LLMs are provided in Section~\ref{sec:app-results-interp-feature}.

%% file: sec/5_discussion.tex
\section{Discussion}
\label{sec:discussion}

Brain alignment for production and comprehension followed similar layerwise trajectories from early layers to those preceding the output layer, with both peaking at intermediate layers.
This result suggests that brain alignment during dialogue is broadly consistent with previous findings on brain alignment during language comprehension~\citep{NEURIPS2019_749a8e6c,Caucheteux2022-hl,antonello2023scaling}.
It is also consistent with previous evidence that some neural activity is shared between production and comprehension~\citep{HASSON2012114,Silbert_2014,Zada2024-to,Yamashita2025-xx}.
At the same time, our results revealed ROI-dependent patterns of brain alignment, such as consistently high alignment across layers in SM for production and in A1 and STG for comprehension.
By examining layerwise brain alignment during dialogue at the whole-brain voxel level using fMRI, our study extends previous findings based on brain activity recorded from a limited number of sEEG channels~\citep{Cai2025-kw}.

We investigated what drives brain alignment during dialogue by examining the dimensionality of LLM representations.
The ID of LLM representations was strongly correlated with whole-cortex brain alignment for both production and comprehension.
This finding suggests that the hypothesis that representational richness contributes to brain alignment, previously proposed for passive language comprehension~\citep{Cheng2026-yc}, also extends to dialogue.
Moreover, this association was strong in regions primarily engaged in each processing mode, whereas it was comparable between the two modes in the language-related regions.
This suggests that the extent to which representational richness accounts for brain activity is relatively independent of processing mode in language-related regions but varies between the two modes in brain regions involved in input and output processing.

Furthermore, we analyzed changes in brain alignment after removing dialogue-related features from LLM representations.
In narrative comprehension, lexical features~\citep{kauf2024lexical,he2025far} and syntactic features~\citep{Oota_2023} represented in LLMs are known to make particularly strong contributions to brain alignment.
In our analysis, lexical features (L) represented in LLMs made the largest contribution to brain alignment, with this tendency being particularly pronounced for production.
This suggests that brain alignment during production may be more strongly influenced by concrete linguistic information, whereas comprehension may additionally involve more abstract, context-dependent information.
The unique contributions of the four features to brain alignment varied across LLM layers and brain regions: BC and L were prominent from early layers onward, primarily in sensorimotor and auditory regions, whereas POS and DA became more prominent from intermediate layers onward in relatively higher-order brain regions.
This pattern is broadly consistent with the functional distinctions between perceptual, motor, language, and higher-order cognitive systems~\citep{fedorenko2024language}.
In particular, \citet{fedorenko2024language} associate the multiple-demand network with frontoparietal cortex and the theory-of-mind network with frontal cortex and the temporoparietal junction.
Accordingly, our finding that DA-related information tends to be dominant in these regions suggests that DA information represented in LLMs may be associated with these networks.

\section{Conclusions}

This study examined the brain alignment of LLMs during dialogue in detail and showed that the contributions of features ranging from low-level to high-level vary across processing modes, LLM layers, and brain regions.
These findings suggest that LLMs can capture aspects of human language and dialogue processing during both production and comprehension. 
They also highlight the value of evaluating the brain alignment of LLMs across multiple processing modes rather than focusing solely on passive language comprehension.

%% file: sec/6_limitations.tex
\section*{Limitations}

We investigated the brain alignment of LLMs during dialogue using three LLMs.
We selected open-source LLMs from different model families and trained on different datasets, and confirmed that they exhibited consistent patterns of brain alignment.
However, whether similar results hold for a wider variety of models and larger LLMs remains to be examined.
Future work with a broader and more diverse set of models will be necessary to determine which findings reflect general properties of LLM representations rather than characteristics specific to the models examined here.

We also investigated the contributions of four interpretable features to the brain alignment of LLMs.
However, these features do not exhaustively characterize the information involved in dialogue.
Although they span multiple levels of representation, from lower-level utterance properties to more abstract dialogue acts, other factors may also contribute to brain alignment.
Extending the set of interpretable features will therefore be important for obtaining a more complete understanding of what drives brain alignment during dialogue.

Our current findings are based on a single fMRI dataset collected during natural dialogue.
Because neural datasets collected during natural dialogue remain scarce, the generalizability of our findings has yet to be established.
Future work should therefore examine whether the brain alignment of LLMs during dialogue reported in this study, as well as the contributions of LLM representational properties to this alignment, can be replicated in other datasets.
Furthermore, the dialogues in the dataset used in this study primarily consist of casual conversations in Japanese about everyday topics.
The communicative goals of real-world dialogue are diverse, including reaching agreement, persuading others, and building interpersonal relationships.
Speakers' behavior, emotional expression may also vary depending on factors such as dialogue topic and cultural contexts.
Future work should therefore investigate brain alignment across a broader range of dialogue settings and communicative contexts.

Finally, the publicly available dialogue data used in this study are limited to text transcripts.
Consequently, our analyses do not incorporate multimodal information such as audio, facial expressions, gaze, and gestures.
Incorporating such multimodal signals may provide a more comprehensive account of the alignment between multimodal LLM representations and human brain activity during dialogue.

%% file: sec/7_ethics.tex
\section*{Ethical Considerations}
\label{sec:ethics}

This study conducts secondary analyses of publicly available fMRI and dialogue data released through OpenNeuro, which were originally collected with approval from the relevant ethics committee.
Because our analyses rely exclusively on publicly available fMRI and dialogue data, the ethical risks associated with this study are minimal.
All analyses were conducted on anonymized data, and no personally identifiable information was accessed or used.

%% file: sec/appendix.tex
\appendix
\renewcommand{\thefigure}{A.\arabic{figure}}
\setcounter{figure}{0}

\section{Additional Methods}
\label{appendix: additional_methods}

\subsection{ROI Selection}
\label{appendix:additional_methods:fmri_data}
We use the publicly available Natural Dialogue fMRI Dataset provided by ~\citet{Yamashita2025-xx}\footnote{\url{https://openneuro.org/datasets/ds004669/versions/2.0.2}}.
We conduct ROI analyses using the eight ROIs provided with the dataset, which are defined based on the anatomical parcellation of the Destrieux Atlas~\cite{destrieux2010automatic}.
Based on previous studies, we select brain regions implicated in language production and comprehension.
As regions involved in semantic and language processing, we select STG, STS, MTG, AG, MFG, and IFG~\cite{binder1997human,chai2016functional}.
We also select SM as a region involved in language production and A1 as a region involved in language comprehension~\cite{Silbert_2014,Goldstein2025-yb, Cai2025-kw, Zada2025-ep}.
Accordingly, we use the following Destrieux Atlas parcels for each ROI: G\_temp\_sup-G\_T\_transv for A1; G\_temp\_sup-Lateral, G\_temp\_sup-Plan\_polar, G\_temp\_sup-Plan\_tempo for STG; S\_temporal\_sup for STS; G\_temporal\_middle for MTG; G\_pariet\_inf-Angular for AG; G\_front\_middle for MFG; G\_front\_inf-Opercular, G\_front\_inf-Orbital, G\_front\_inf-Triangul for IFG; G\_precentral, S\_central, G\_postcentral for SM.

\subsection{Large Language Models}
\label{appendix:additional_methods:llms}

This study uses three language models available on Hugging Face: tokyotech-llm/Llama-3.1-Swallow-8B-Instruct-v0.5\footnote{\url{https://huggingface.co/tokyotech-llm/Llama-3.1-Swallow-8B-Instruct-v0.5}}, tokyotech-llm/Gemma-2-Llama-Swallow-9b-it-v0.1\footnote{\url{https://huggingface.co/tokyotech-llm/Gemma-2-Llama-Swallow-9b-it-v0.1}}, and tokyotech-llm/GPT-OSS-Swallow-20B-SFT-v0.1\footnote{\url{https://huggingface.co/tokyotech-llm/GPT-OSS-Swallow-20B-SFT-v0.1}}, hereafter referred to in this section as Llama-3.1-Swallow, Gemma-2-Swallow, and GPT-OSS-Swallow, respectively.

Llama-3.1-Swallow and Gemma-2-Swallow were developed by further instruction-tuning Llama-3.1-8B-Instruct ~\citep{grattafiori2024llama3herdmodels} and gemma-2-9b-it~\citep{gemmateam2024gemma2improvingopen}, respectively, on a Japanese-language corpora.
GPT-OSS-Swallow was developed through continued pretraining and supervised fine-tuning of gpt-oss-20b~\citep{openai2025gptoss120bgptoss20bmodel} on a Japanese-English corpus.

Llama-3.1-Swallow has 32 layers, a hidden dimension of 4,096, approximately 8B parameters, and a vocabulary size of 128,256.
Gemma-2-Swallow has 42 layers, a hidden dimension of 3,584, approximately 9B parameters, and a vocabulary size of 256,000.
Both models employ dense architectures.
In contrast, GPT-OSS-Swallow employs a mixture-of-experts (MoE) architecture, with 24 layers, a hidden dimension of 2,880 and a vocabulary size of 201,088.
It has approximately 21B parameters in total, of which 3.6B are active per token.

\subsection{Head-Motion Correction}
\label{appendix:additional_methods:headmotion}
To account for head-motion-related variance in the BOLD signal, we use the head-motion feature constructed by \citet{Yamashita2025-xx}.
We construct voxel-wise encoding models to predict fMRI activity from this feature, using the same regression and cross-validation procedures described in Section~\ref{sec:methods-alignment}.
We then subtract the predicted activity from the original fMRI activity and use the resulting residuals for subsequent brain alignment analyses.

\subsection{Estimation of Intrinsic Dimensions}
\label{appendix:additional_methods:IDs}

We estimate IDs using GRIDE~\citep{Denti2022}, which generalizes the TwoNN estimator~\citep{Facco2017-hk} to multiple neighborhood scales.
GRIDE is based on the following ratio of nearest-neighbor distances:
\begin{equation}
  \mu_{i,k,2k} = \frac{r_{i,2k}}{r_{i,k}}\notag
\end{equation}
where \(r_{i,j}\) represents the Euclidean distance between point \(i\) and its \(j\)-th nearest neighbor. 
Under the assumption of a locally uniform density distribution, the ratios \(\mu_{i,k,2k}\) follow a generalized Pareto distribution:
\begin{equation}
  f_{\mu_{i,k,2k}}(\mu) 
  = \frac{d \left(\mu^{\,d}-1\right)^{k-1}}{B(k,k)\mu^{\,d(2k-1)+1}}\notag
\end{equation}
where \(B(\cdot,\cdot)\) is the beta function. By further assuming that the ratios \(\mu_{i,k,2k}\) are independent across data points, the ID can be estimated by numerically maximizing the likelihood of this distribution.

For each LLM and participant, we select a scale \(k\) at which the estimated ID is most stable.
First, we estimate the ID for each LLM layer at multiple scales \(k=2^s\).
Let \(d_{l,s}\) denote the estimated ID at layer \(l\) and scale \(k=2^s\).
For each candidate scale, we calculate the relative change in ID using a central difference:
\begin{equation}
  \delta_{l,s}
  =
  \frac{|d_{l,s+1}-d_{l,s-1}|}
  {2\max(|d_{l,s}|,\varepsilon)}
  \notag
\end{equation}
where \(\varepsilon=10^{-8}\) prevents division by zero.
We then define the stability score for each scale as the median relative change across all LLM layers:
\begin{equation}
  S_s = \operatorname{median}(\delta_{1,s},\ldots,\delta_{L,s})
  \notag
\end{equation}
Finally, we select the scale that minimizes the stability score:
\begin{equation}
  s^* = \underset{s\in\mathcal{S}}{\arg\min}\; S_s,
  \qquad k^*=2^{s^*}
  \notag
\end{equation}
where \(\mathcal{S}\) denotes the set of candidate scales indices, restricted to \(s\leq 8\) and excluding the smallest and largest evaluated scales.

\subsection{Interpretable Features Related to Dialogue}
\label{appendix:additional_methods:interpretable_features}

\paragraph{Bunsetsu Count}
The bunsetsu count feature (BC) represents the number of bunsetsu in each TR-level text segment.
The count is categorized into six groups (1, 2, 3, 4, 5, and 6 or more), yielding a six-dimensional one-hot vector.
We treat this as a surface-level linguistic feature related to utterance length.

\paragraph{Lexicon}
The lexicon feature (L) represents the presence or absence of each word in a vocabulary list. 
We constructed a vocabulary list from the Corpus of Everyday Japanese Conversation (CEJC)~\cite{koiso-etal-2022-design} to focus on words commonly used in everyday conversation.
We selected the 1,000 most frequent words from the CEJC vocabulary list (ver. 2022.09).
For each TR-level text segment, we represent the presence or absence of each word's lemma from this list as a binary value (1/0), yielding a 1,000-dimensional multi-hot vector.
We treat L as a surface-level linguistic feature that captures lexical information.

\paragraph{Part of Speech}
The part-of-speech feature (POS) represents the presence or absence of tokens belonging to each of nine POS categories: nouns, verbs, adjectives, adverbs, auxiliary verbs, conjunctions, pronouns, interjections, and particles.
This yields a nine-dimensional multi-hot vector.
We treat this as a syntactic feature.
For the construction of the BC, L, and POS features, we used the GiNZA NLP Library\footnote{\url{https://github.com/megagonlabs/ginza}} for Japanese bunsetsu segmentation and tokenization.

\paragraph{Dialogue Act}
Dialogue acts (DAs) refer to the communicative intentions underlying speakers' utterances in dialogue.
The DA feature is constructed by annotating TR-level text segments with DA tags related to everyday conversation.
We use 10 DA tags selected from a DA tag set designed for speakers and listeners in non-task-oriented dialogue~\cite{meguro-etal-2010-controlling}.
Table~\ref{tab:annotation-da} presents the DA tags and their definitions.
DA annotation was performed manually by two annotators.
The annotators were instructed to assign the most appropriate tag to each TR-level text segment based on the definitions of the 10 DA tags, considering both the utterance and its surrounding context.
To ensure consistency between the two annotators, we assign a value of 1 to a DA tag only when both annotators agree on that tag, and 0 otherwise.
The resulting binary values for the 10 DA tags are concatenated to form a 10-dimensional multi-hot vector for each TR-level text segment.
We use the resulting feature matrix as the DA features.
We treat these as features related to communicative intentions.

\section{Additional Results}
\subsection{Brain Alignment of LLMs during Dialogue}
\label{sec:app-results-brain-alignment}

\begin{figure*}[t]
\centering
\includegraphics[width=0.95\linewidth]{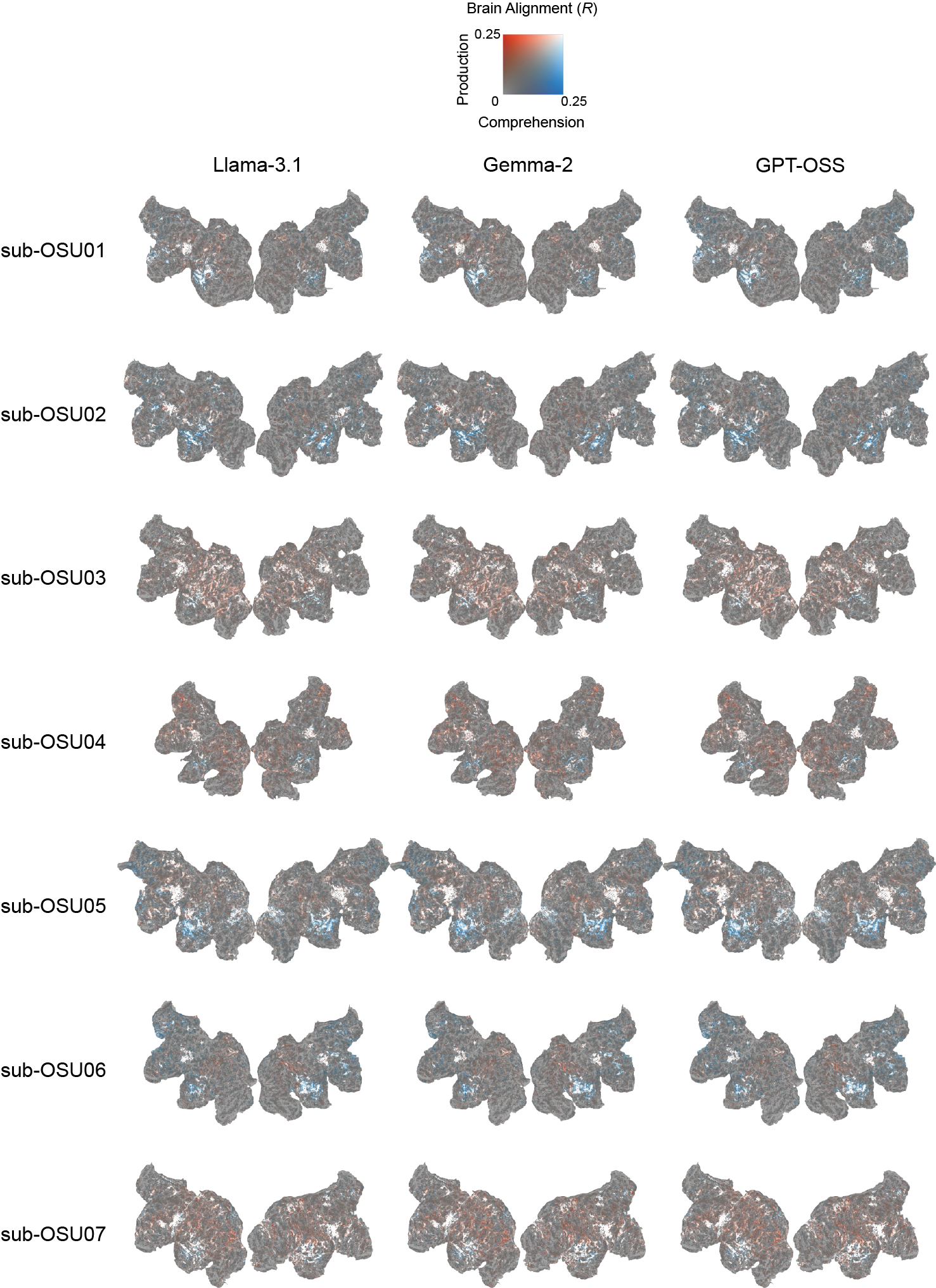}
\caption{\textbf{Voxelwise Brain Alignment of Llama-3.1, Gemma-2 and GPT-OSS during Dialogue.} . 
}
\label{fig:App1}
\end{figure*}

\begin{figure*}[t]
\centering
\includegraphics[width=0.95\linewidth]{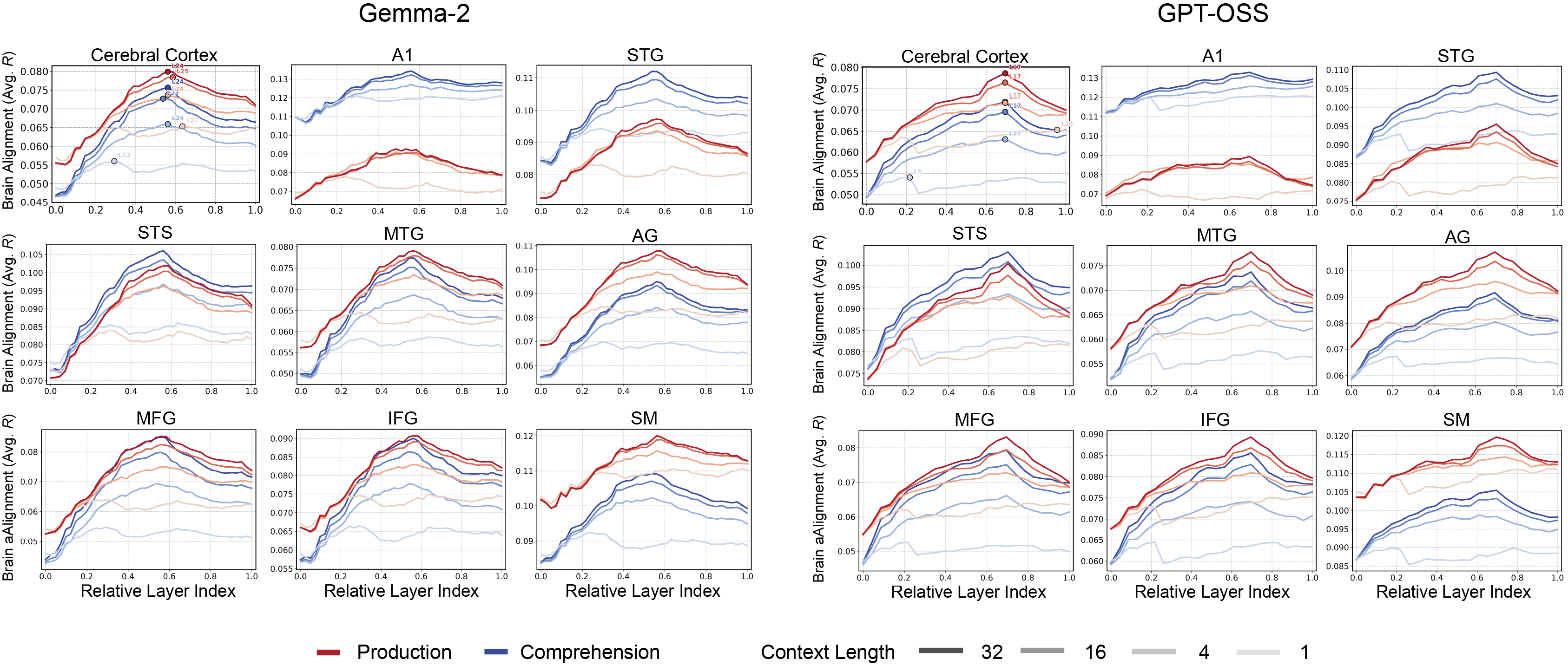}
\caption{\textbf{Brain Alignment of Gemma-2 and GPT-OSS during Dialogue.} 
Mean brain alignment across the cerebral cortex (top left) and within each ROI (remaining panels) for production and comprehension across Gemma-2 (left) and GPT-OSS (right) layers, averaged across participants. Color intensity indicates context length, and circles indicate the peak layer for each condition. 
}
\label{fig:App2}
\end{figure*}

\begin{figure*}[t]
\centering
\includegraphics[width=1\linewidth]{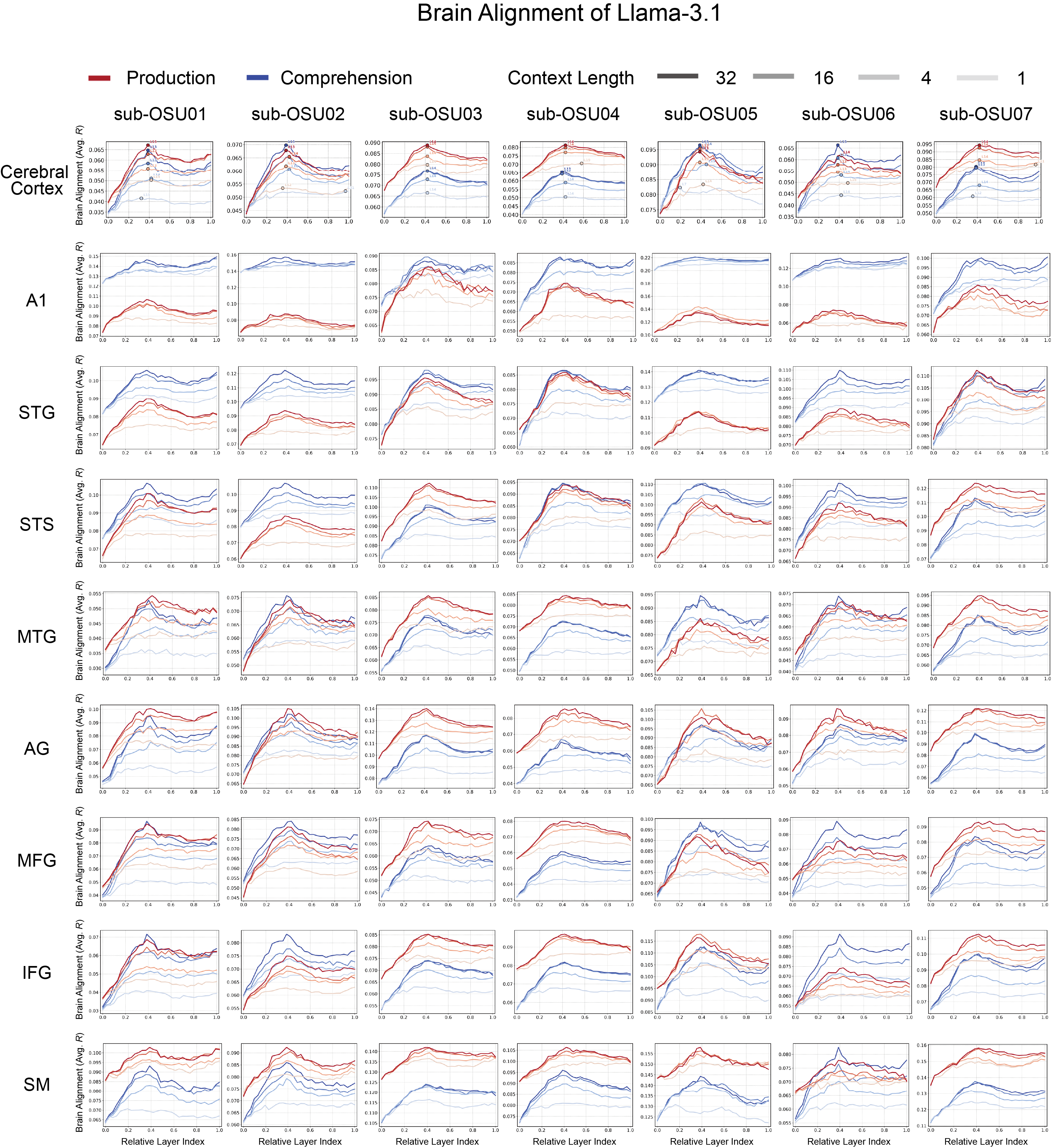}
\caption{\textbf{Brain Alignment of Llama-3.1 during Dialogue for Each Participant.} 
Mean brain alignment for each participant across the cerebral cortex (top) and within each ROI (remaining panels) for production and comprehension across Llama-3.1 layers. Color intensity indicates context length, and circles indicate the peak layer for each condition. 
}
\label{fig:App3_llama_alignment}
\end{figure*}

\begin{figure*}[t]
\centering
\includegraphics[width=1\linewidth]{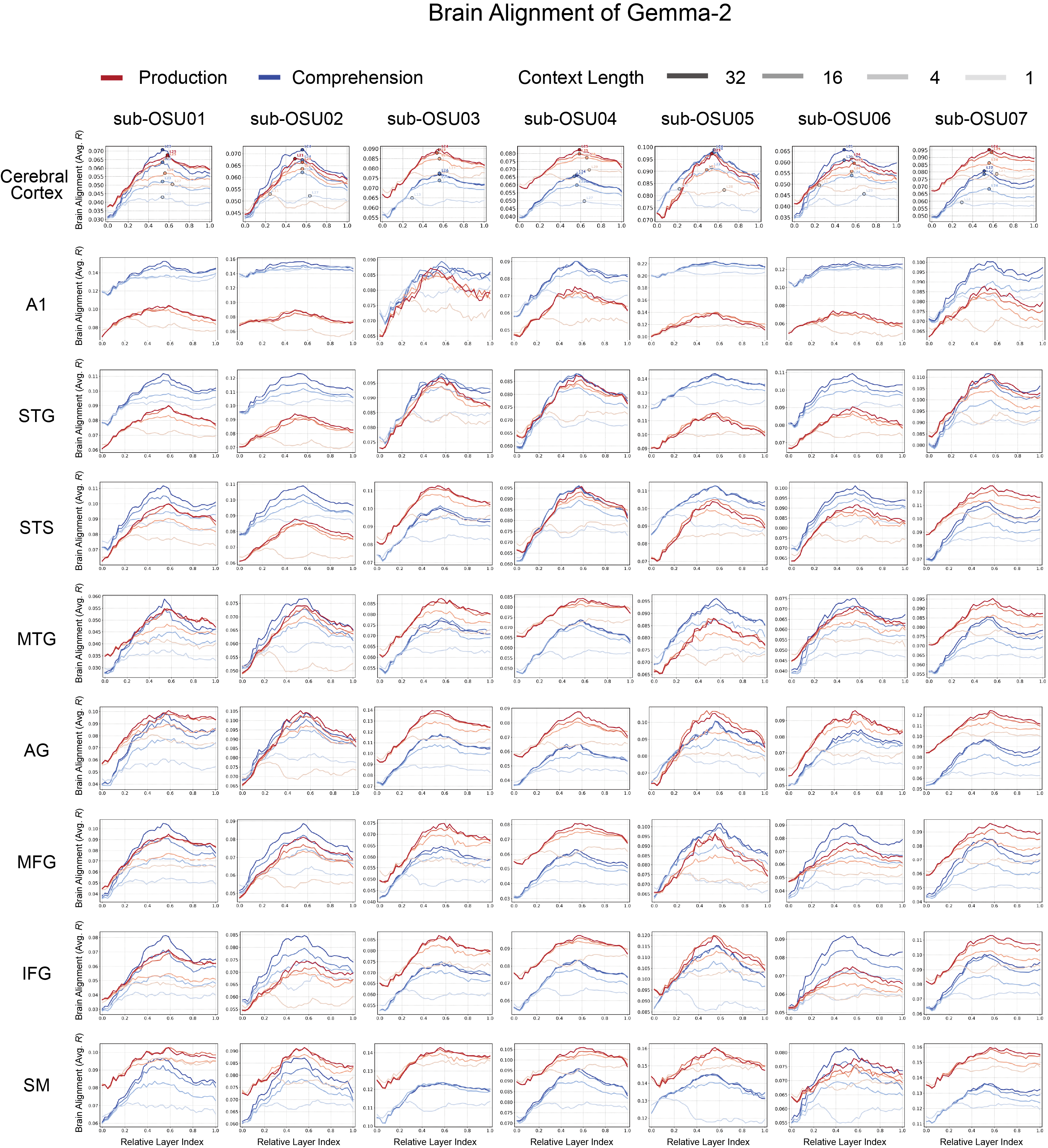}
\caption{\textbf{Brain Alignment of Gemma-2 during Dialogue for Each Participant.} 
Mean brain alignment for each participant across the cerebral cortex (top) and within each ROI (remaining panels) for production and comprehension across Gemma-2 layers. Color intensity indicates context length, and circles indicate the peak layer for each condition. 
}
\label{fig:App4_gemma_alignment}
\end{figure*}

\begin{figure*}[t]
\centering
\includegraphics[width=1\linewidth]{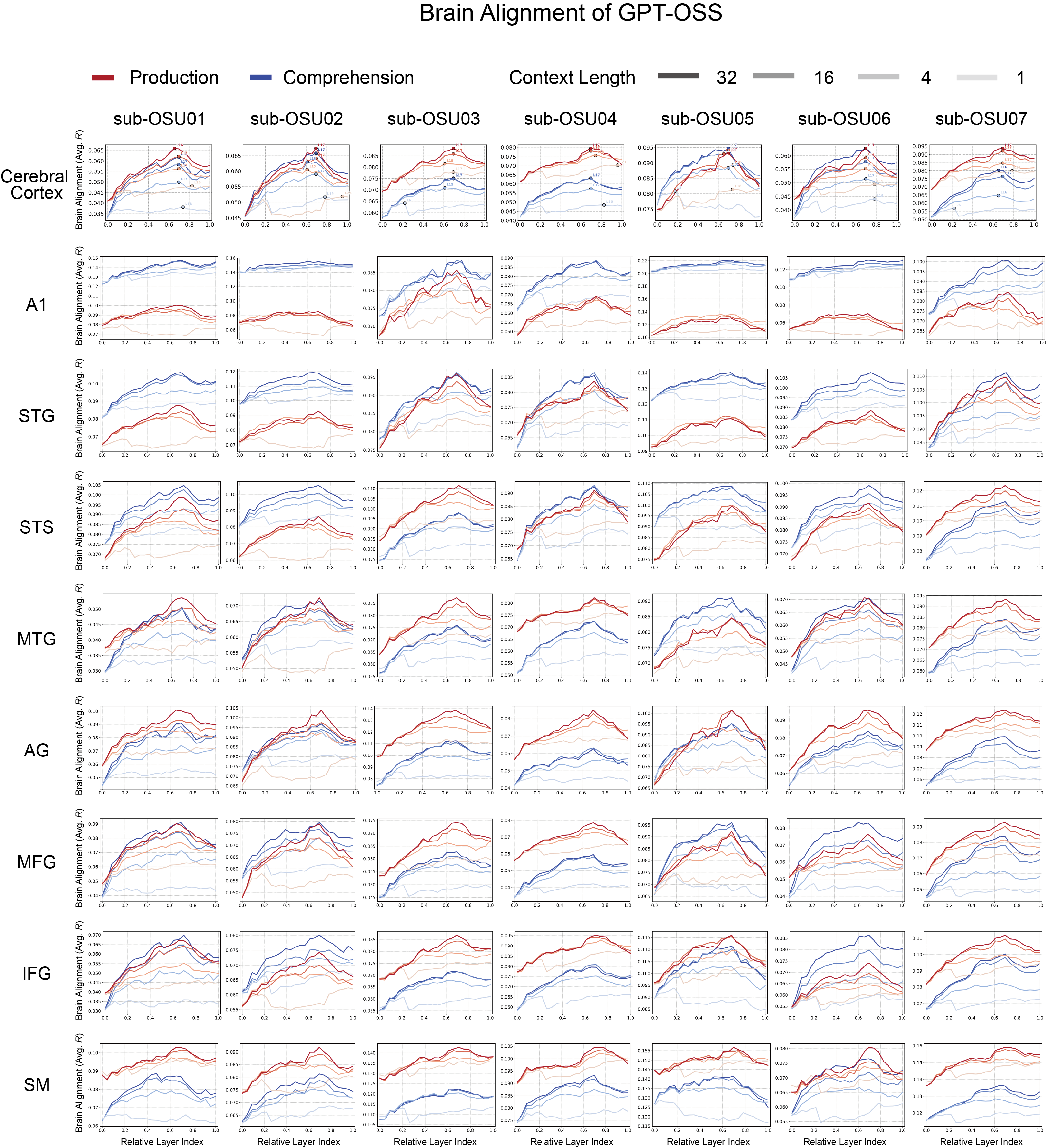}
\caption{\textbf{Brain Alignment of GPT-OSS during Dialogue for Each Participant.} 
Mean brain alignment for each participant across the cerebral cortex (top) and within each ROI (remaining panels) for production and comprehension across GPT-OSS layers. Color intensity indicates context length, and circles indicate the peak layer for each condition. 
}
\label{fig:App5_gptoss_alignment}
\end{figure*}

This section presents supplementary results related to Figure~\ref{fig:prod_vs_comp} in Section~\ref{sec:results-brain-alignment}.
Figure~\ref{fig:App1} shows the results corresponding to Figure~\ref{fig:prod_vs_comp}a for all participants and all three LLMs, revealing similar spatial patterns of brain alignment across all conditions.
Figure~\ref{fig:App2} shows the results corresponding to Figure~\ref{fig:prod_vs_comp}b for the other two LLMs, Gemma-2 and GPT-OSS.
Across all three LLMs, brain alignment consistently peaks at intermediate layers, despite differences in the number of layers.
Furthermore, all three LLMs consistently exhibit high brain alignment across layers in SM and AG for production and in A1 and STG for comprehension.
In addition, brain alignment in A1 for comprehension increases again near the output layers across all three LLMs.
These results demonstrate that the layerwise patterns of cortex-averaged and ROI-averaged brain alignment for production and comprehension are consistent across all three LLMs.
Figures~\ref{fig:App3_llama_alignment},~\ref{fig:App4_gemma_alignment}, and~\ref{fig:App5_gptoss_alignment} present the corresponding participant-level results for Figures~\ref{fig:prod_vs_comp}b and~\ref{fig:App2}, showing similar patterns across individual participants.

\subsection{Probing Performance of Interpretable Features}
\label{sec:app-probing-interp-feature}

This section presents the probing results for the interpretable features described in Section~\ref{sec:methods-comp-commint}.
Table~\ref{tab:probing_before_after_ridge} shows the probing performance evaluated using ridge regression.
For both production and comprehension, the correlation coefficients decrease substantially after feature removal across all features.
These results suggest that the feature removal procedure operates as intended, substantially reducing the amount of linearly decodable information about the targeted features in LLM representations.

\subsection{Effects of Interpretable Features Related to Dialogue}
\label{sec:app-results-interp-feature}

\begin{figure*}[t]
\centering
\includegraphics[width=1\linewidth]{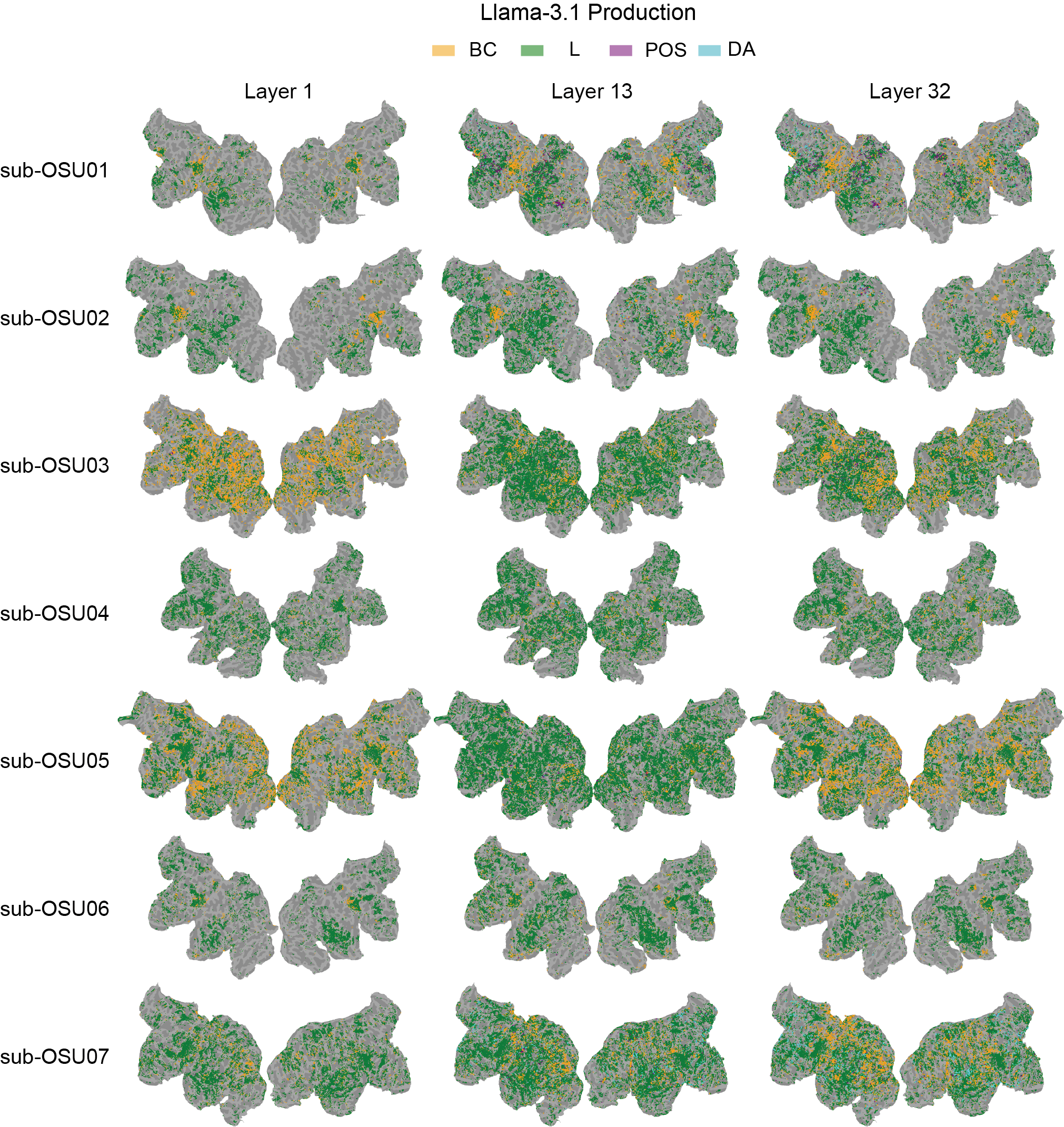}
\caption{\textbf{Distribution of Dominant Interpretable Dialogue-Related Features for Brain Alignment during Production in Llama-3.1.} For Layers 1, 13, and 32 of Llama-3.1, the feature whose removal caused the largest decrease in brain alignment at each voxel is mapped onto the cerebral cortex of each participant. Only voxels with significant brain alignment ($P < 0.05$, FDR corrected) of at least 0.1 in the full model are colored.}
\label{fig:App7_interp_feat_llama_prod}
\end{figure*}

\begin{figure*}[t]
\centering
\includegraphics[width=1\linewidth]{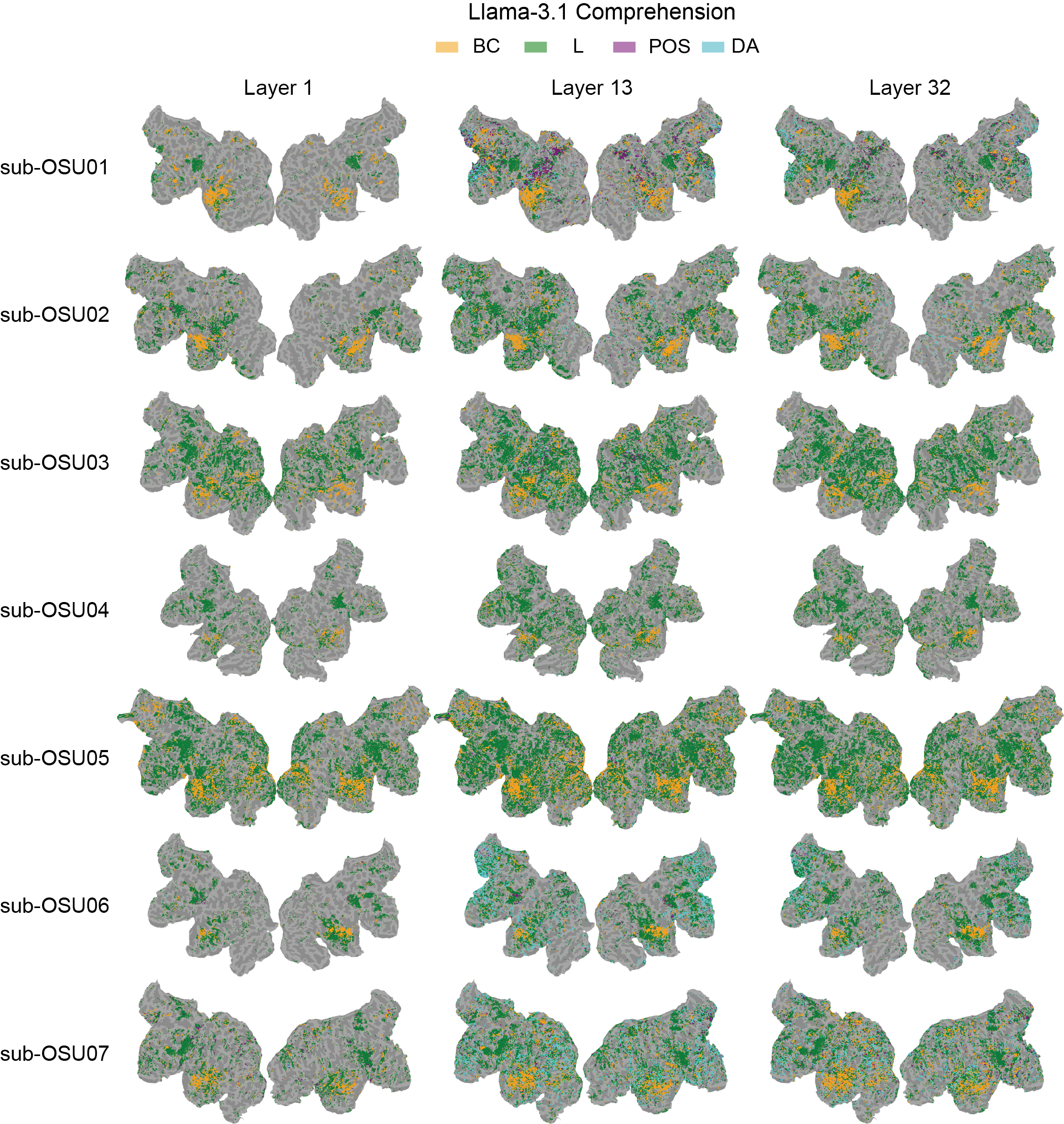}
\caption{\textbf{Distribution of Dominant Interpretable Dialogue-Related Features for Brain Alignment during Comprehension in Llama-3.1.} For Layers 1, 13, and 32 of Llama-3.1, the feature whose removal caused the largest decrease in brain alignment at each voxel is mapped onto the cerebral cortex of each participant. Only voxels with significant brain alignment ($P < 0.05$, FDR corrected) of at least 0.1 in the full model are colored.}
\label{fig:App8_interp_feat_llama_comp}
\end{figure*}

\begin{figure*}[t]
\centering
\includegraphics[width=1\linewidth]{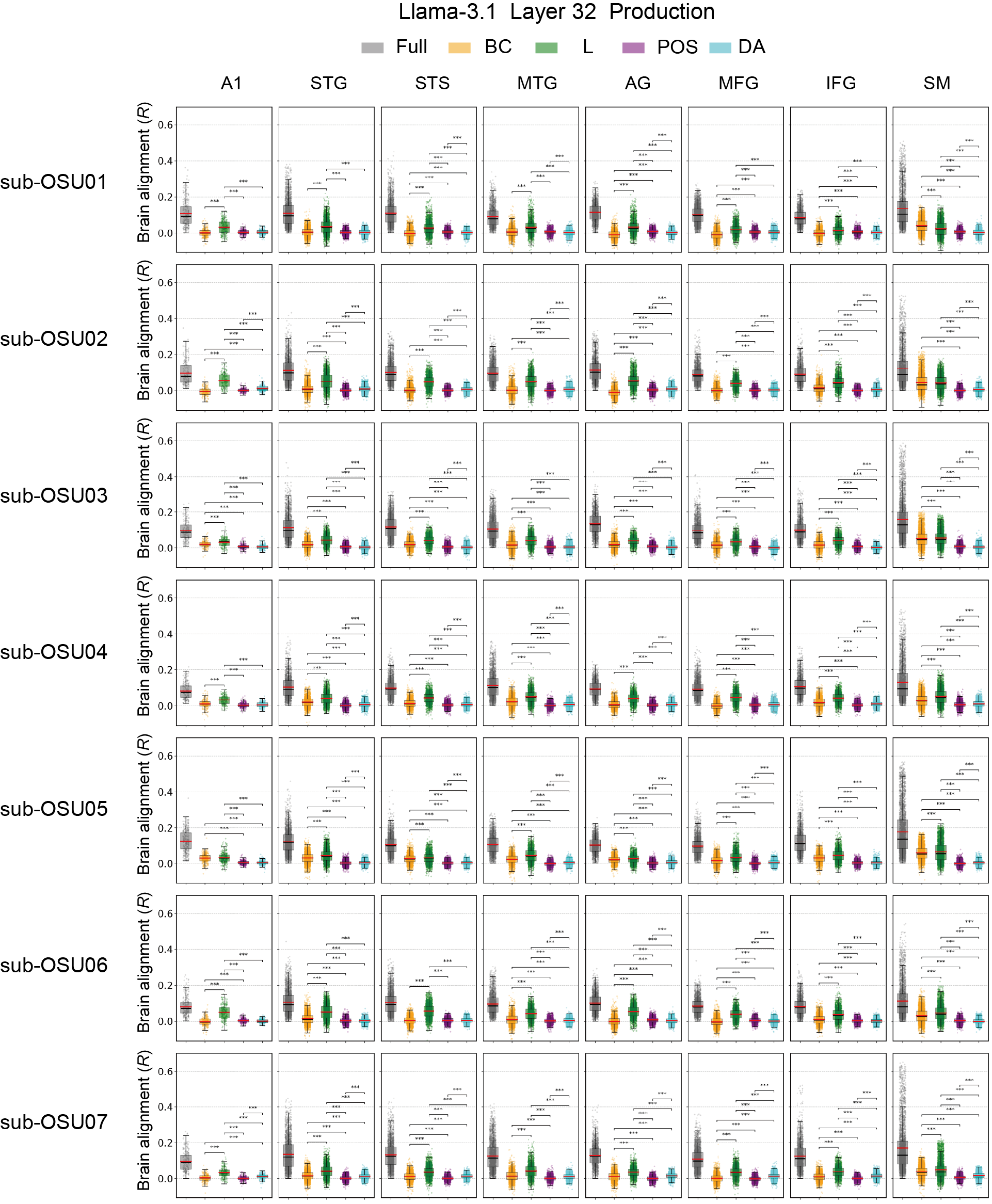}
\caption{\textbf{Unique Contributions of Interpretable Dialogue-Related Features to Brain Alignment of Layer 32 of Llama-3.1 during Production.} Boxes indicate the median and interquartile range, red lines indicate the mean, and points represent individual voxels. Gray indicates brain alignment in the full model, while yellow, green, purple, and blue indicate the unique contributions of BC, L, POS, and DA, respectively. Asterisks mark significant differences between features ($P < 0.001$, two-sided Wilcoxon signed-rank tests with Holm correction).}
\label{fig:App9_interp_feat_llama_l32_prod}
\end{figure*}

\begin{figure*}[t]
\centering
\includegraphics[width=1\linewidth]{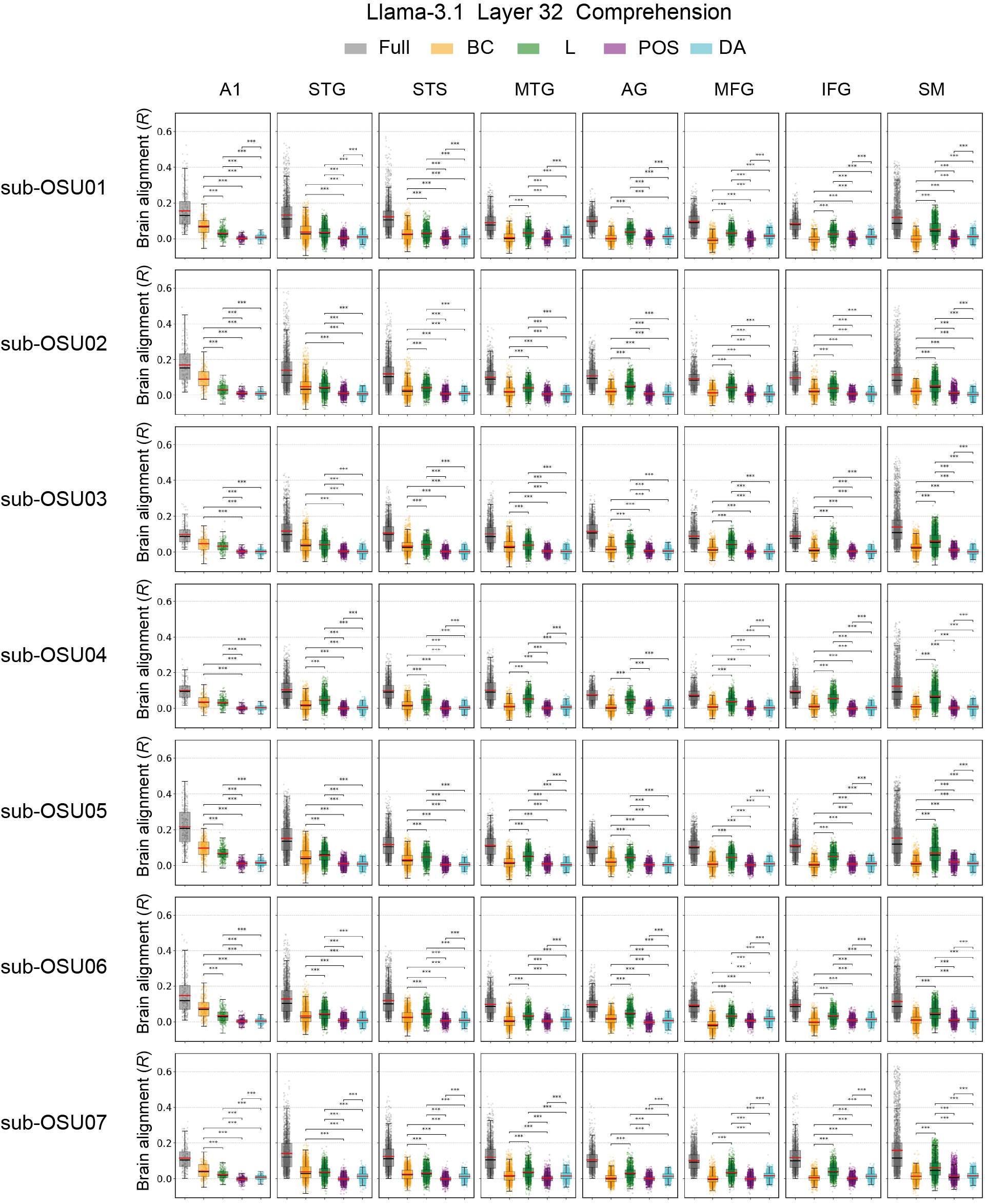}
\caption{\textbf{Unique Contributions of Interpretable Dialogue-Related Features to Brain Alignment of Layer 32 of Llama-3.1 during Comprehension.} Boxes indicate the median and interquartile range, red lines indicate the mean, and points represent individual voxels. Gray indicates brain alignment in the full model, while yellow, green, purple, and blue indicate the unique contributions of BC, L, POS, and DA, respectively. Asterisks mark significant differences between features ($P < 0.001$, two-sided Wilcoxon signed-rank tests with Holm correction).}
\label{fig:App10_interp_feat_llama_l32_comp}
\end{figure*}

\begin{figure*}[t]
\centering
\includegraphics[width=1\linewidth]{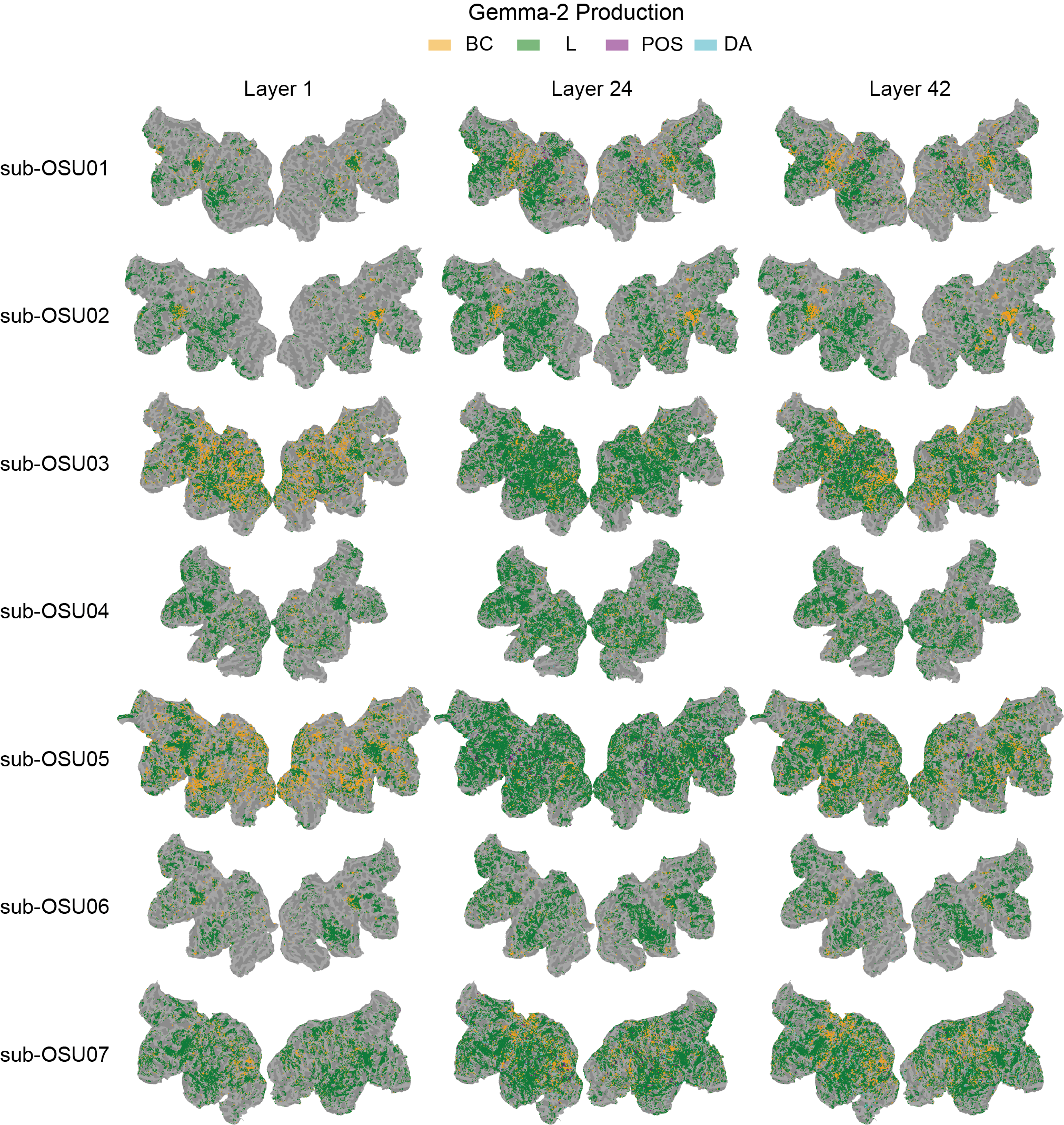}
\caption{\textbf{Distribution of Dominant Interpretable Dialogue-Related Features for Brain Alignment during Production in Gemma-2.} For Layers 1, 24, and 42 of Gemma-2, the feature whose removal caused the largest decrease in brain alignment at each voxel is mapped onto the cerebral cortex of each participant. Only voxels with significant brain alignment ($P < 0.05$, FDR corrected) of at least 0.1 in the full model are colored.}
\label{fig:App11_interp_feat_gemma_prod}
\end{figure*}

\begin{figure*}[t]
\centering
\includegraphics[width=1\linewidth]{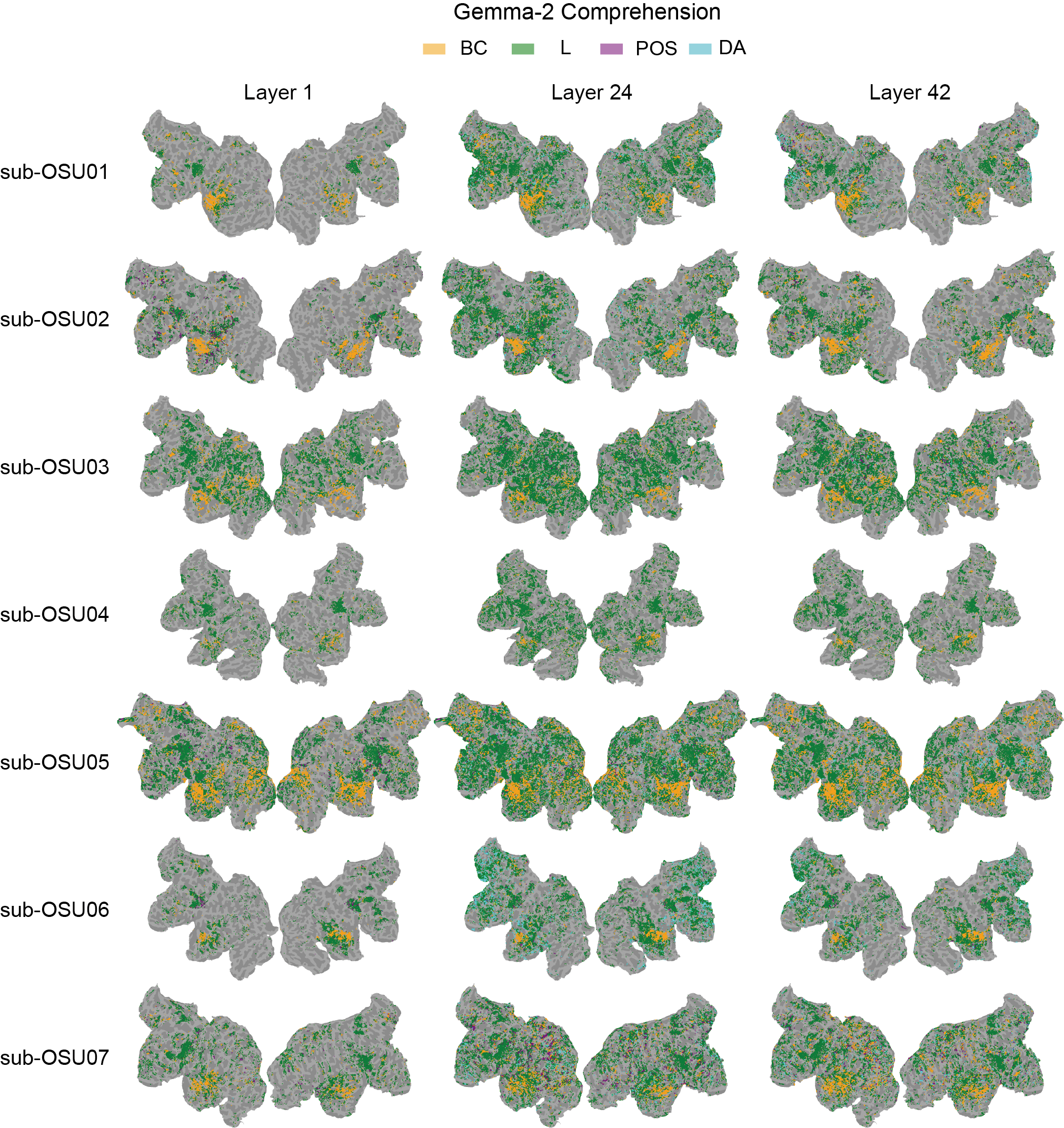}
\caption{\textbf{Distribution of Dominant Interpretable Dialogue-Related Features for Brain Alignment during Comprehension in Gemma-2.} For Layers 1, 24, and 42 of Gemma-2, the feature whose removal caused the largest decrease in brain alignment at each voxel is mapped onto the cerebral cortex of each participant. Only voxels with significant brain alignment ($P < 0.05$, FDR corrected) of at least 0.1 in the full model are colored.}
\label{fig:App12_interp_feat_gemma_comp}
\end{figure*}

\begin{figure*}[t]
\centering
\includegraphics[width=1\linewidth]{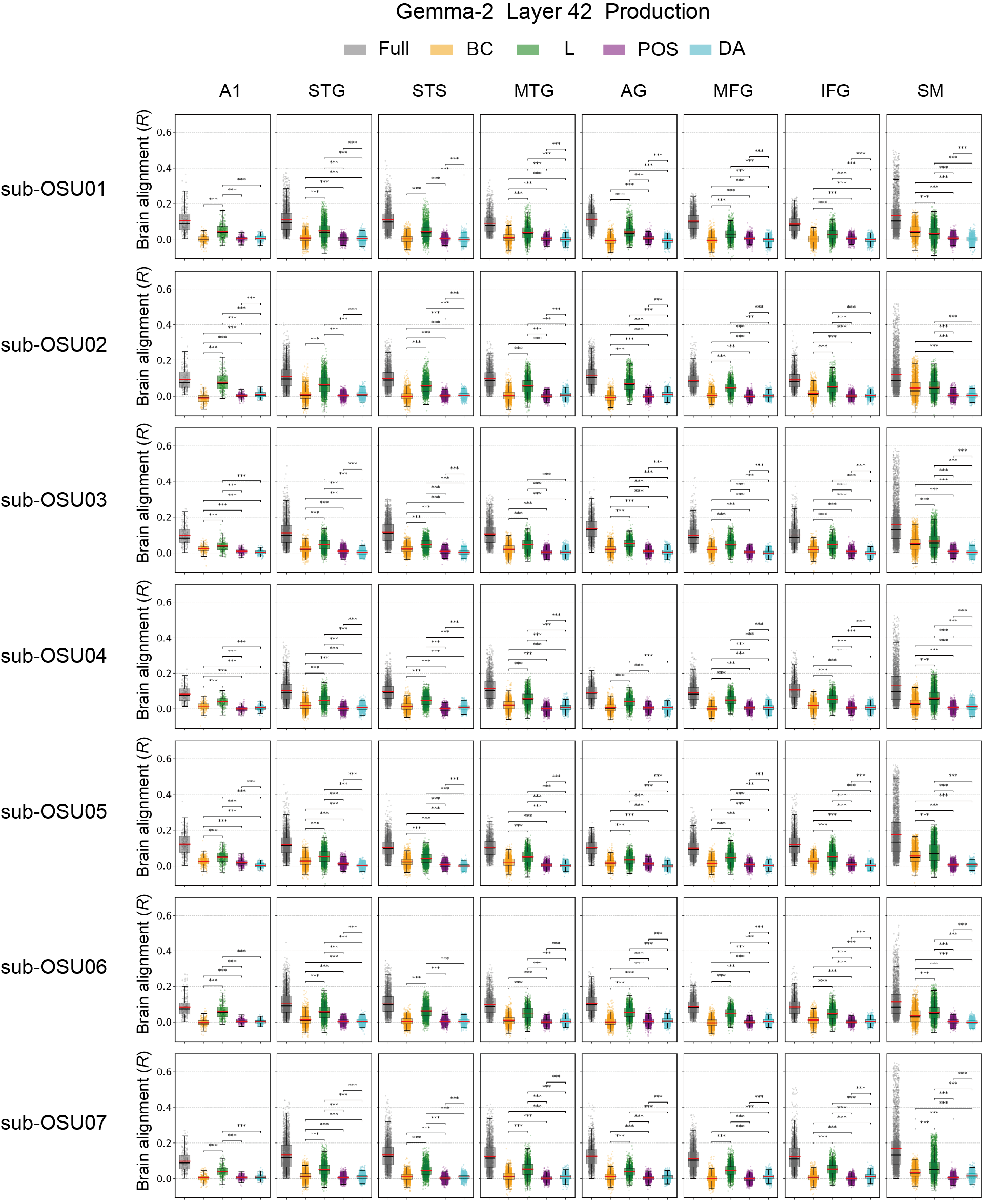}
\caption{\textbf{Unique Contributions of Interpretable Dialogue-Related Features to Brain Alignment of Layer 42 of Gemma-2 during Production.} Boxes indicate the median and interquartile range, red lines indicate the mean, and points represent individual voxels. Gray indicates brain alignment in the full model, while yellow, green, purple, and blue indicate the unique contributions of BC, L, POS, and DA, respectively. Asterisks mark significant differences between features ($P < 0.001$, two-sided Wilcoxon signed-rank tests with Holm correction).}
\label{fig:App13_interp_feat_gemma_l42_prod}
\end{figure*}

\begin{figure*}[t]
\centering
\includegraphics[width=1\linewidth]{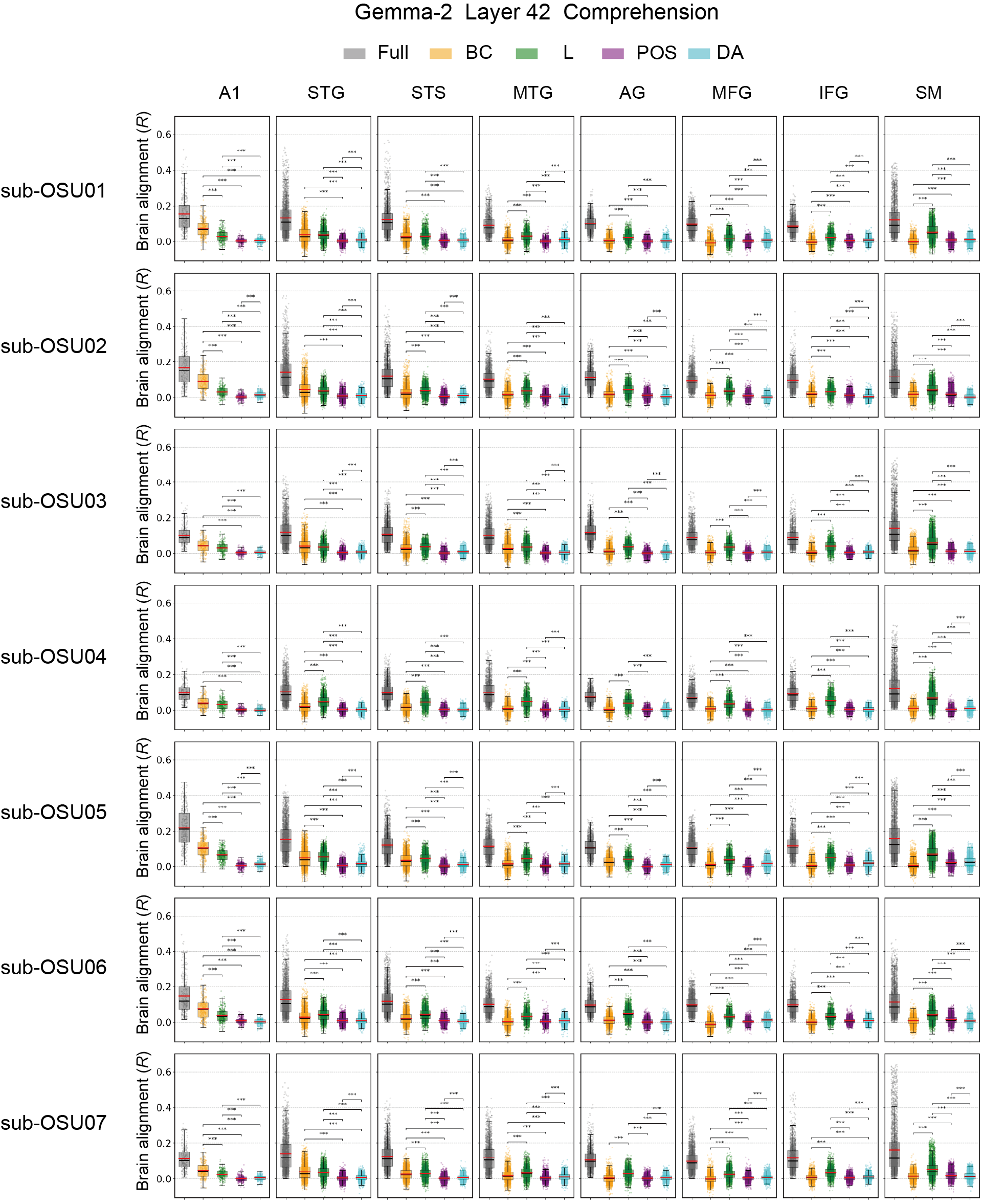}
\caption{\textbf{Unique Contributions of Interpretable Dialogue-Related Features to Brain Alignment of Layer 42 of Gemma-2 during Comprehension.} Boxes indicate the median and interquartile range, red lines indicate the mean, and points represent individual voxels. Gray indicates brain alignment in the full model, while yellow, green, purple, and blue indicate the unique contributions of BC, L, POS, and DA, respectively. Asterisks mark significant differences between features ($P < 0.001$, two-sided Wilcoxon signed-rank tests with Holm correction).}
\label{fig:App14_interp_feat_gemma_l42_comp}
\end{figure*}

\begin{figure*}[t]
\centering
\includegraphics[width=1\linewidth]{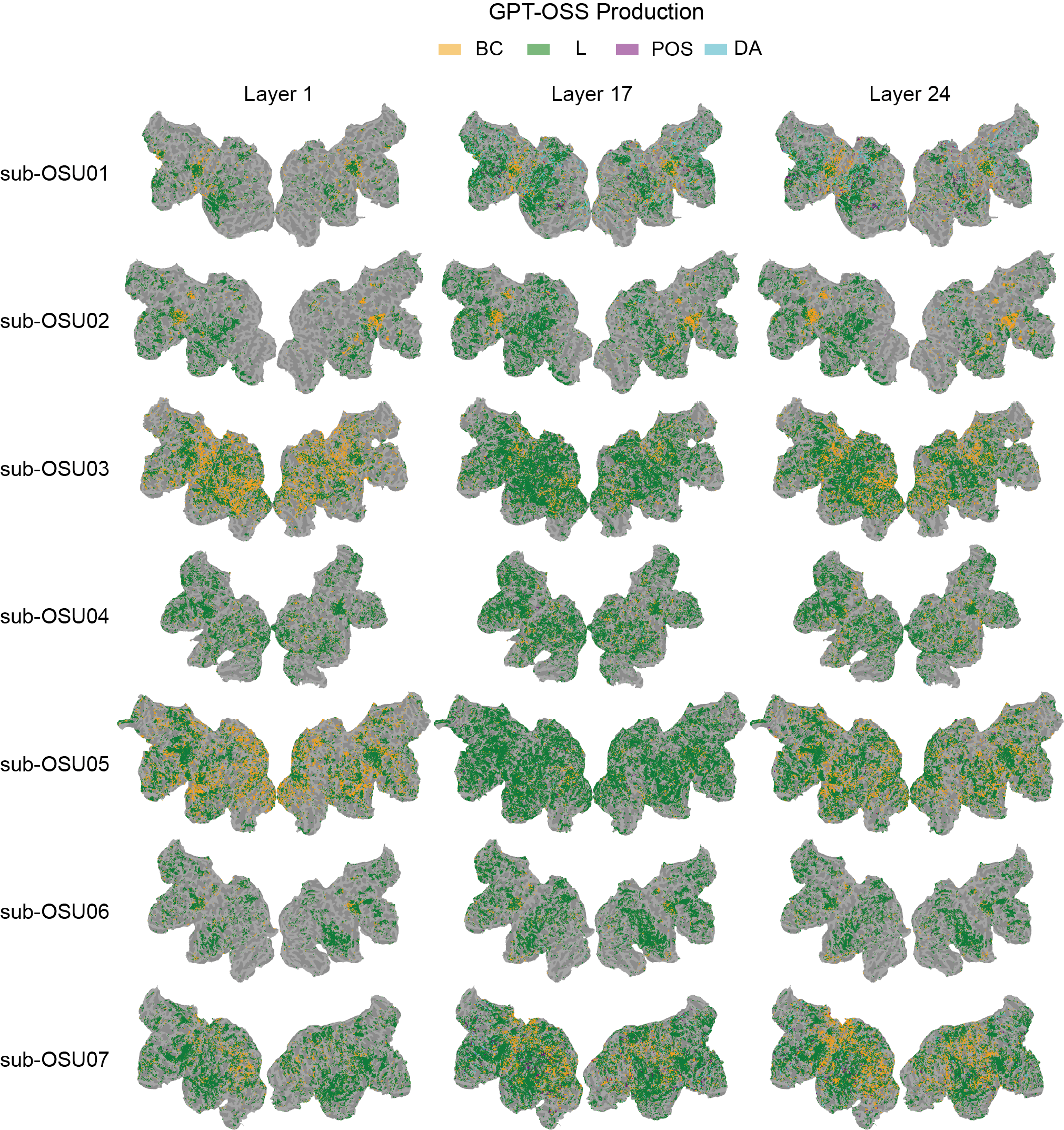}
\caption{\textbf{Distribution of Dominant Interpretable Dialogue-Related Features for Brain Alignment during Production in GPT-OSS.} For Layers 1, 17, and 24 of GPT-OSS, the feature whose removal caused the largest decrease in brain alignment at each voxel is mapped onto the cerebral cortex of each participant. Only voxels with significant brain alignment ($P < 0.05$, FDR corrected) of at least 0.1 in the full model are colored.}
\label{fig:App15_interp_feat_gptoss_prod}
\end{figure*}

\begin{figure*}[t]
\centering
\includegraphics[width=1\linewidth]{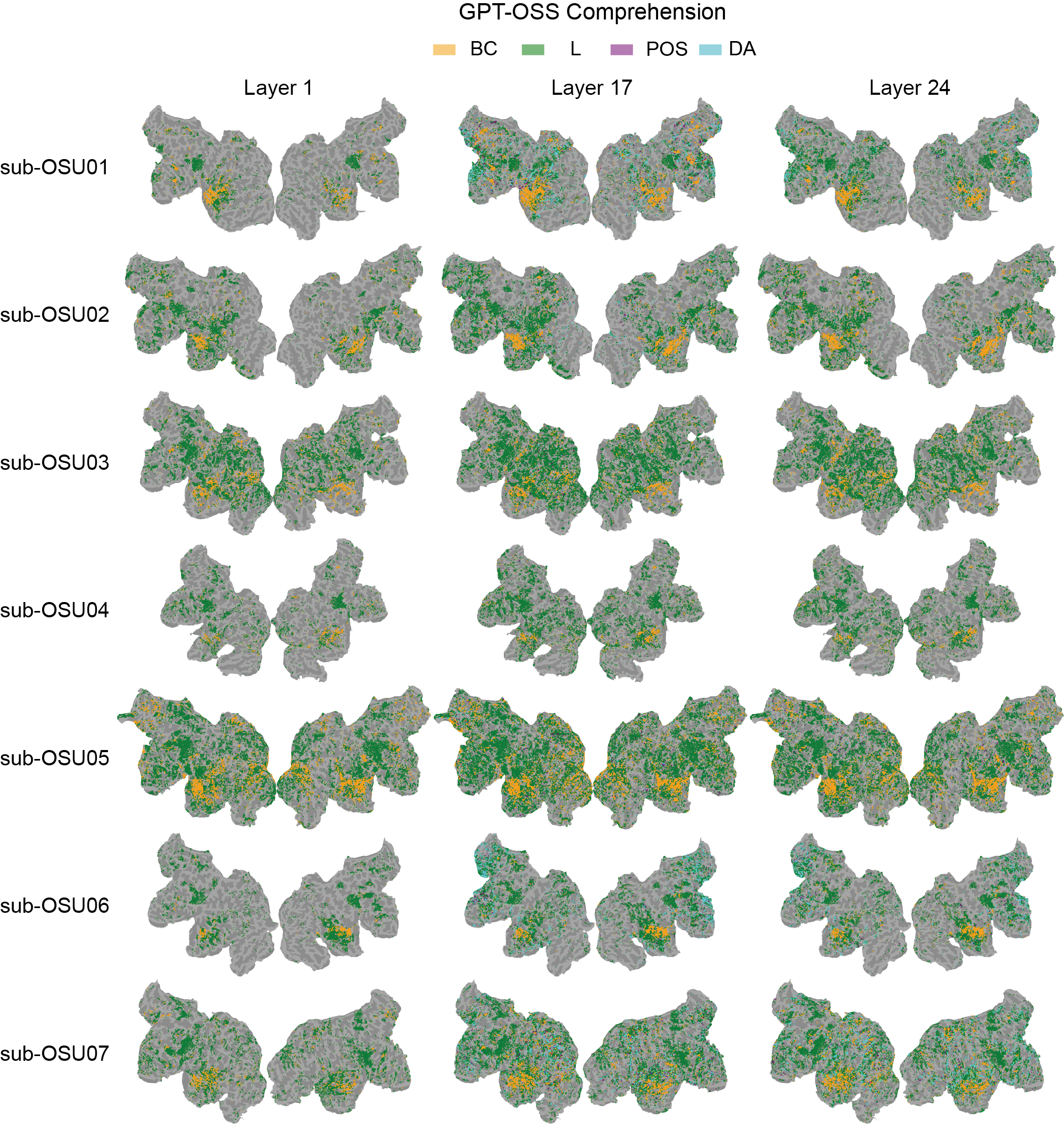}
\caption{\textbf{Distribution of Dominant Interpretable Dialogue-Related Features for Brain Alignment during Comprehension in GPT-OSS.} For Layers 1, 17, and 24 of GPT-OSS, the feature whose removal caused the largest decrease in brain alignment at each voxel is mapped onto the cerebral cortex of each participant. Only voxels with significant brain alignment ($P < 0.05$, FDR corrected) of at least 0.1 in the full model are colored.}
\label{fig:App16_interp_feat_gptoss_comp}
\end{figure*}

\begin{figure*}[t]
\centering
\includegraphics[width=1\linewidth]{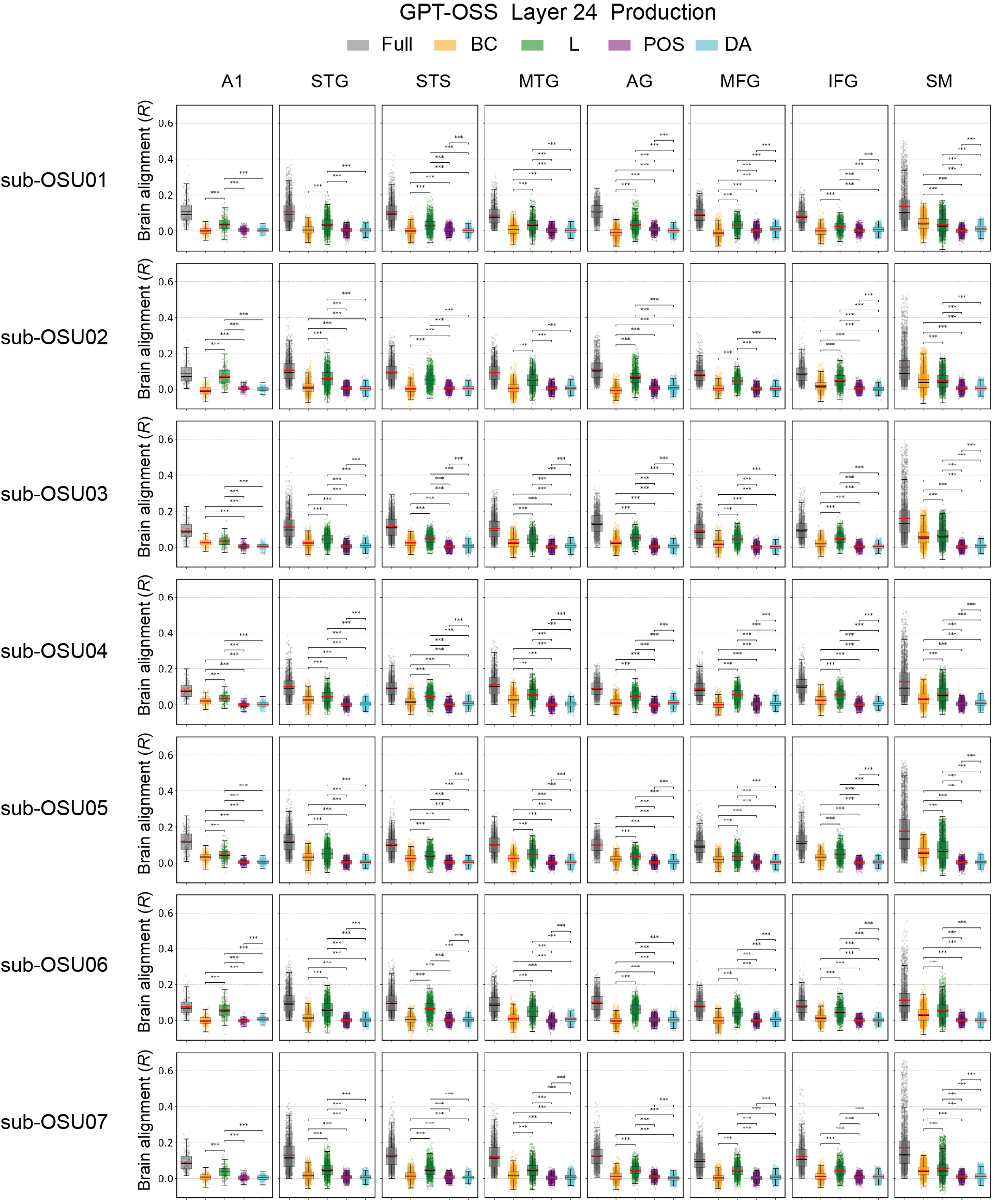}
\caption{\textbf{Unique Contributions of Interpretable Dialogue-Related Features to Brain Alignment of Layer 24 of GPT-OSS during Production.} Boxes indicate the median and interquartile range, red lines indicate the mean, and points represent individual voxels. Gray indicates brain alignment in the full model, while yellow, green, purple, and blue indicate the unique contributions of BC, L, POS, and DA, respectively. Asterisks mark significant differences between features ($P < 0.001$, two-sided Wilcoxon signed-rank tests with Holm correction).}
\label{fig:App17_interp_feat_gptoss_l24_prod}
\end{figure*}

\begin{figure*}[t]
\centering
\includegraphics[width=1\linewidth]{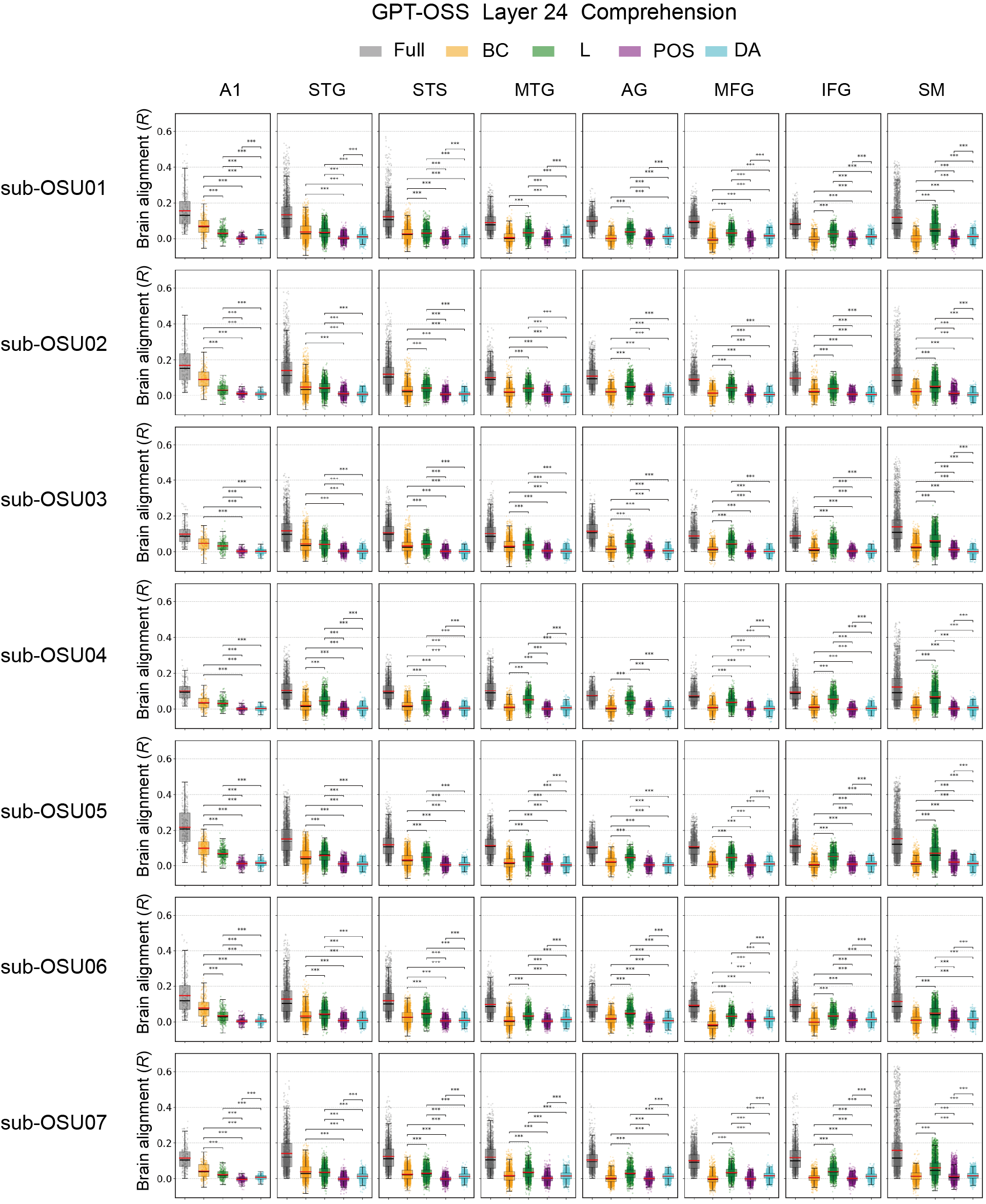}
\caption{\textbf{Unique Contributions of Interpretable Dialogue-Related Features to Brain Alignment of Layer 24 of GPT-OSS during Comprehension.} Boxes indicate the median and interquartile range, red lines indicate the mean, and points represent individual voxels. Gray indicates brain alignment in the full model, while yellow, green, purple, and blue indicate the unique contributions of BC, L, POS, and DA, respectively. Asterisks mark significant differences between features ($P < 0.001$, two-sided Wilcoxon signed-rank tests with Holm correction).}
\label{fig:App18_interp_feat_gptoss_l24_comp}
\end{figure*}

This section presents supplementary results related to Section~\ref{sec:results-Interp_feature}.
Tables~\ref{tab:gemma_layer_results} and~\ref{tab:gptoss_layer_results} present the results corresponding to Table~\ref{tab:llama_layer_results} for the other two LLMs, Gemma-2 and GPT-OSS.
Consistent with the results for Llama-3.1, L makes the largest contribution to brain alignment across layers for both production and comprehension in Gemma-2 and GPT-OSS, with this tendency being particularly pronounced for production.
Figures~\ref{fig:App7_interp_feat_llama_prod},~\ref{fig:App8_interp_feat_llama_comp},~\ref{fig:App11_interp_feat_gemma_prod},~\ref{fig:App12_interp_feat_gemma_comp},~\ref{fig:App15_interp_feat_gptoss_prod}, and~\ref{fig:App16_interp_feat_gptoss_comp} present the flatmap results corresponding to Figure~\ref{fig:effect-interp-feat} for all participants and all three LLMs.
Similar spatial patterns of dominant features are observed across participants and LLMs, particularly for BC and L.
In addition, for comprehension, POS and DA also emerge as dominant features in voxels within temporoparietal, parietal, and frontal regions from intermediate layers onward, further supporting the consistency of our findings.
Figures~\ref{fig:App9_interp_feat_llama_l32_prod},~\ref{fig:App10_interp_feat_llama_l32_comp},~\ref{fig:App13_interp_feat_gemma_l42_prod},~\ref{fig:App14_interp_feat_gemma_l42_comp},~\ref{fig:App17_interp_feat_gptoss_l24_prod}, and~\ref{fig:App18_interp_feat_gptoss_l24_comp} present the boxplot results corresponding to Figure~\ref{fig:effect-interp-feat} for all participants and all three LLMs.
Consistent with the flatmap results, the relative magnitudes of the unique contributions of each feature across ROIs are broadly consistent across participants and LLMs.

\begin{table*}[t]
\small
\centering
\renewcommand{\arraystretch}{1.15}
\caption{Dialogue act tags and their definitions.}
\begin{tabular}{@{}p{0.2\textwidth}p{0.78\textwidth}@{}}
\toprule
Dialogue Act & Definition \\
\midrule
INFORMATION & Utterances that provide objective information or convey information obtained from others, including hearsay. \\
SELF-DISCLOSURE & Utterances in which the speaker expresses personal information, opinions, or feelings. \\
BACKCHANNEL & Utterances that indicate that the speaker is listening or encourage the interlocutor to continue speaking. \\
QUESTION & Utterances in which the speaker asks the interlocutor for information or a response. \\
SYMPATHY & Utterances that express agreement with the interlocutor. \\
NON-SYMPATHY & Utterances that reject the interlocutor's opinion or express an opposing view. \\
REPEAT & Utterances that repeat the interlocutor's words or content, either verbatim or through paraphrasing. \\
FILLER & Utterances produced while the speaker is thinking about what to say next. \\
ADMIRATION & Utterances that express interest or emotion through interjections such as ``Ah.'' \\
OTHER & Utterances that do not belong to any of the above categories. \\
\bottomrule
\end{tabular}
\label{tab:annotation-da}
\end{table*}

\begin{figure*}[t]
\centering
\includegraphics[width=0.8\linewidth]{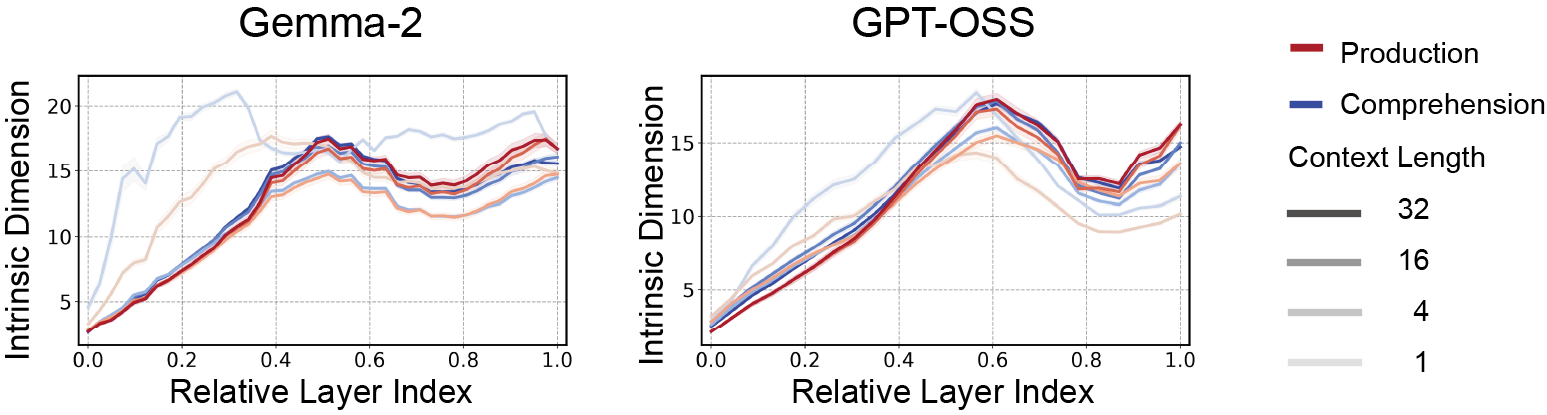}
\caption{\textbf{Relationship between LLM representational complexity and brain alignment during dialogue.} Layerwise ID of Gemma-2 (left) and GPT-OSS (right) representations for production and comprehension, averaged across participants' dialogue. Shaded regions indicate the standard error of the mean. Red and blue indicate production and comprehension, respectively, and color intensity indicates context length.}
\label{fig:App6_id_plot}
\end{figure*}

\begin{table*}[t]
\caption{Comparison of ridge regression probing performance (Pearson's $r$) before and after removing interpretable features across LLM layers for production and comprehension.}
\centering
\scriptsize
\setlength{\tabcolsep}{3pt}
\begin{tabular}{ll*{16}{c}}
\toprule
\multirow{3}{*}{Models} & \multirow{3}{*}{Layers}
& \multicolumn{8}{c}{Production}
& \multicolumn{8}{c}{Comprehension} \\
\cmidrule(lr){3-10} \cmidrule(lr){11-18}
& & \multicolumn{2}{c}{BC} & \multicolumn{2}{c}{L}
  & \multicolumn{2}{c}{POS} & \multicolumn{2}{c}{DA}
  & \multicolumn{2}{c}{BC} & \multicolumn{2}{c}{L}
  & \multicolumn{2}{c}{POS} & \multicolumn{2}{c}{DA} \\
& & Before & After & Before & After & Before & After & Before & After
  & Before & After & Before & After & Before & After & Before & After \\
\midrule
\multirow{4}{*}{Llama-3.1}
& 1  & .38 & -.00 & .33 & .01 & .72 & .01 & .41 & .01
     & .36 & -.02 & .32 & .01 & .70 & .01 & .41 & .01 \\
& 13 & .43 & -.01 & .30 & .01 & .75 & .01 & .51 & .01
     & .41 & -.03 & .29 & .01 & .74 & .02 & .51 & .00 \\
& 26 & .34 & -.00 & .26 & .01 & .71 & .03 & .48 & .00
     & .34 & -.03 & .26 & .01 & .71 & .03 & .48 & .00 \\
& 32 & .28 & .01 & .23 & .01 & .68 & .05 & .45 & .02
     & .29 & -.02 & .23 & .01 & .68 & .05 & .46 & .01 \\
\midrule
\multirow{4}{*}{Gemma-2}
& 1  & .38 & -.02 & .34 & .02 & .71 & .00 & .39 & .02
     & .36 & -.03 & .31 & .02 & .69 & .01 & .37 & .01 \\
& 12 & .48 & -.00 & .42 & .02 & .80 & .02 & .50 & .01
     & .45 & -.02 & .39 & .02 & .78 & .02 & .49 & .02 \\
& 24 & .41 & -.01 & .28 & .02 & .74 & .04 & .51 & .01
     & .41 & -.02 & .27 & .02 & .73 & .03 & .51 & .01 \\
& 42 & .27 & .02 & .24 & .01 & .68 & .07 & .45 & .03
     & .28 & -.01 & .24 & .01 & .68 & .07 & .46 & .03 \\
\midrule
\multirow{4}{*}{GPT-OSS}
& 1  & .39 & .02 & .34 & .11 & .74 & .08 & .42 & .06
     & .37 & .02 & .33 & .11 & .72 & .10 & .42 & .07 \\
& 8  & .44 & .01 & .35 & .02 & .77 & .03 & .49 & .04
     & .41 & -.01 & .33 & .02 & .76 & .03 & .48 & .04 \\
& 17 & .40 & .00 & .27 & .01 & .73 & .03 & .51 & .02
     & .39 & -.02 & .27 & .01 & .72 & .02 & .50 & .00 \\
& 24 & .27 & .03 & .23 & .01 & .67 & .06 & .44 & .03
     & .28 & -.01 & .23 & .01 & .67 & .06 & .45 & .02 \\
\bottomrule
\end{tabular}
\label{tab:probing_before_after_ridge}
\end{table*}

\begin{table}[t]
\caption{\textbf{Mean Percentage Decrease in Brain Alignment after Removing Interpretable Features.} Percentage decrease ($\%$) in whole-cortex-averaged brain alignment of Gemma-2 after removing each of the four feature types. Results are averaged across participants and shown for layers 1, 12, 24, and 42 for both production (Prod) and comprehension (Comp).}
\centering
\small
\begin{tabular}{llrrrr}
\toprule
 & & L1 & L12 & L24 & L42 \\
\midrule
\multirow{4}{*}{Prod}
& BC  & 30.64 & 13.69 & 7.87 & 12.78 \\
& L  & \textbf{86.34} & \textbf{66.38} & \textbf{48.52} & \textbf{55.75} \\
& POS & 66.97 & 45.44 & 28.87 & 38.76 \\
& DA  & 67.35 & 40.80 & 25.72 & 35.56 \\
\midrule
\multirow{4}{*}{Comp}
& BC  & 27.67 & 12.84 & 8.45 & 10.36 \\
& L  & \textbf{78.60} & \textbf{56.47} & \textbf{41.19} & \textbf{45.90} \\
& POS & 74.46 & 44.48 & 28.13 & 37.64 \\
& DA  & 78.71 & 47.93 & 32.55 & 41.37 \\
\bottomrule
\end{tabular}
\label{tab:gemma_layer_results}
\end{table}

\begin{table}[t]
\caption{\textbf{Mean Percentage Decrease in Brain Alignment after Removing Interpretable Features.} Percentage decrease ($\%$) in whole-cortex-averaged brain alignment of GPT-OSS after removing each of the four feature types. Results are averaged across participants and shown for layers 1, 8, 17, and 24 for both production (Prod) and comprehension (Comp).}
\centering
\small
\begin{tabular}{llrrrr}
\toprule
 & & L1 & L8 & L17 & L24 \\
\midrule
\multirow{4}{*}{Prod}
& BC  & 33.38 & 15.72  & 9.87 & 15.61 \\
& L  & \textbf{89.16} & \textbf{69.60} & \textbf{52.24} & \textbf{58.88} \\
& POS & 69.67 & 45.53 & 31.82 & 41.85 \\
& DA  & 72.14 & 42.11 & 30.40 & 37.53 \\
\midrule
\multirow{4}{*}{Comp}
& BC  & 29.17 & 13.77  & 9.68  & 9.87 \\
& L  & \textbf{85.86} & \textbf{56.78} & \textbf{43.18} & \textbf{50.15} \\
& POS & 75.59 & 46.00 & 33.36 & 36.59 \\
& DA  & 79.37 & 49.28 & 34.90 & 42.18 \\
\bottomrule
\end{tabular}
\label{tab:gptoss_layer_results}
\end{table}

